\documentclass{ieeeaccess}
\usepackage{cite}
\usepackage{amsmath,amssymb,amsfonts}
\usepackage{algorithmic}
\usepackage{enumitem}
\usepackage{graphicx}
\usepackage{textcomp}
\usepackage[hidelinks]{hyperref} 
\usepackage{tabularx}
\usepackage{float} 
\usepackage[table]{xcolor}
\usepackage{amsmath,amssymb,amsfonts}
\usepackage{rotating}
\usepackage{algorithmic}
\usepackage{xcolor}   
\usepackage{textcomp}
 \usepackage{booktabs}
 \usepackage{booktabs}
\usepackage{pdflscape}
\usepackage{caption}
\usepackage{amssymb}   
\usepackage{pifont}   
\usepackage{adjustbox}
\newcommand{\cmark}{\checkmark}    
\newcommand{\xmark}{\ding{55}}      
\newcommand{\partialmark}{\textopenbullet}
 \usepackage{stfloats}
\def\BibTeX{{\rm B\kern-.05em{\sc i\kern-.025em b}\kern-.08em
T\kern-.1667em\lower.7ex\hbox{E}\kern-.125emX}}
\usepackage{array}
\vol{13}
\year{2025}
\begin{document}
\setcounter{page}{177647}
\history{Date of publication October 6, 2025.}
\doi{10.1109/ACCESS.2025.3618109}

\title{Systematic Literature Review of Machine Learning Models and Applications for Text Recognition}
\author{\uppercase{NUZHAT KHAN}\authorrefmark{1}, \uppercase{AB AL-HADI AB RAHMAN}\authorrefmark{1} \IEEEmembership{Senior Member, IEEE}, 
\uppercase{Shahriyar Masud Rizvi}\authorrefmark{3}, \uppercase{Ibrahim Yousef Alshareef}\authorrefmark{1}, \uppercase{MUHAMMAD NADZIR MARSONO}\authorrefmark{1}, \uppercase{MuhAMMAD PAEND BAKHT}\authorrefmark{4}
\IEEEmembership{Member, IEEE},
\uppercase{MOHD SHAHRIZAL RUSLI}\authorrefmark{2},
and \uppercase{SHAHIDATUL SADIAH}\authorrefmark{1}}

\address[1]{Faculty of Electrical Engineering, Universiti Teknologi Malaysia,  Johor Bahru 81310, Malaysia (khan.nuzhat@utm.my), (hadi@utm.my), (Yousefm@graduate.utm.my), (shahidatulsadiah@utm.my), (mnadzir@utm.my) }
\address[2]{Faculty of Artificial Intelligence, Universiti Teknologi Malaysia, Kuala Lumpur 54100, Malaysia (shahrizalr@utm.my)}
\address[3]{Department of Electrical and Electronic Engineering, American International University-Bangladesh, 408/1, Kuratoli 1229, Bangladesh (shahriyar@aiub.edu)}
\address[4]{Department of Electrical Engineering, Balochistan University of Information Technology, Engineering and Management Sciences, Quetta 87300, Pakistan (paend.bakht@gmail.com)}
\tfootnote{This work was supported in part by Universiti Teknologi Malaysia with the Flagship CoE/RG research grant number Q.J130000.5023.10G05 and Q.J130000.5023.10G09, and the Professional Development grant number R.J130000.7113.07E51. This work is licensed under a Creative Commons Attribution 4.0 License. For more information, see \url{https://creativecommons.org/licenses/by/4.0/}.}

\markboth
{Khan \headeretal: Systematic Literature Review of Machine Learning Models and Applications for Text Recognition}
{Khan \headeretal: Systematic Literature Review of Machine Learning Models and Applications for Text Recognition}

\corresp{Corresponding author: Ab Al-Hadi Ab Rahman (hadi@utm.my).}

\begin{abstract}
Optical Character Recognition (OCR) for text recognition using machine vision has
significantly improved, particularly when handling heterogeneous textual data. Traditional
OCR models struggle with script variations, writing styles, and degraded documents. 
 Advancements in technology are leading to new AI models with improved architecture for handling multiple languages and complex data formats. Despite this progress, a comprehensive evaluation of OCR advancements remains limited. Based on the established preferred reporting items for systematic reviews and meta-analysis (PRISMA) guidelines, this literature review presents an extensive assessment of OCR research to trace the evolution of AI models over the past decade. It explores the transition in AI models, application domains, data types, linguistic coverage,
and challenges. Through a detailed analysis of 97 selected studies published during January 2015 - January 2025, key OCR
models are identified, and their performance, strengths, and limitations are analyzed.
The findings highlight how OCR technologies have evolved to address structured and unstructured text, scene text recognition, and multilingual processing. Unresolved challenges include limited resources for underrepresented languages, high variability in handwritten text, visual similarity among characters, and constraints in real-time OCR applications. To address these issues, several promising approaches are proposed. Key suggestions include self-supervised learning, multimodal AI, automated machine learning (AutoML), AI-assisted postprocessing, tiny machine learning (TinyML), and the creation of joint corpora for script matching. The
future recommendations aim to enhance OCR accuracy and tackle the challenges identified for real-time
industrial applications. This study will guide future research and establish a foundation for OCR
field.
\end{abstract}


\titlepgskip=-15pt

\maketitle

\section{Introduction}
\label{sec:introduction}
\begin{figure}[!t]
 \hspace{-2cm}
  \centering
 
\includegraphics[width=1\columnwidth]{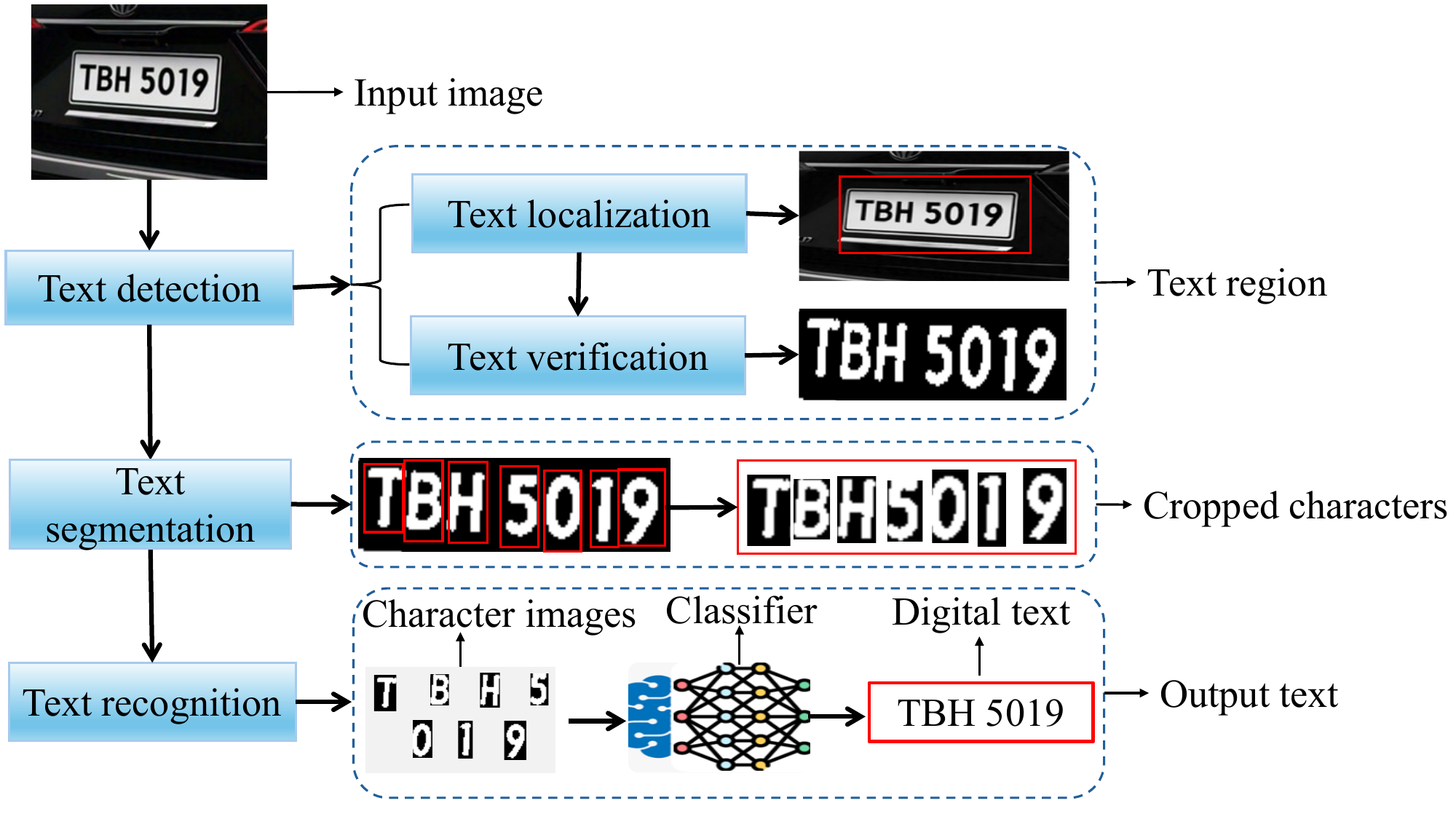} 
  \caption{Example of end-to-end car plate OCR.}
\label{fig:figure1}
\end{figure}

\PARstart{O}{ptical} character recognition (OCR) systems have transformed the way we interact with textual information \cite{abhishek2020optical}. Recent advancements in artificial intelligence (AI) and machine vision have significantly enhanced the ability of computers to interpret and process visual information. Now, AI models analyze their surroundings through image processing to extract meaningful insights. One key application of image processing is text extraction, where AI models use OCR to convert printed or handwritten text into a machine-readable format \cite{abdullah2023systematic}. See a simplified illustration of an end-to-end OCR system for a car license plate recognition in \autoref{fig:figure1}. The process begins with text detection, followed by text segmentation, and concludes with text recognition. Associated techniques, such as text localization and text verification, complement the main process. These dynamic and problem-specific techniques may include multiple steps of image processing, classification, and natural language processing \cite{thorat2022detailed}.
Initially developed to digitize printed and handwritten documents, OCR has evolved into a standard tool for modern data management and the digital preservation of cultural heritage \cite{drobac2020optical}. Today, OCR technology is widely adopted across various industries for seamless human-computer interaction \cite{kopitar2025review}. In the finance sector, OCR streamlines check processing and fraud detection \cite{khandokar2024computer}. In healthcare, it facilitates the digitization of patient records and the automation of medical billing \cite{karthikeyan2021ocr}. The education sector benefits from the digitization of books and research papers, which increases access to academic resources \cite{naseer2024investigating}. In the legal field, OCR supports contract analysis and digitization of case files \cite{baviskar2021efficient} while government agencies use it to digitize public records for efficient services\cite{ruangraweenukit2024autonomous,singh2024review,imam2022ocr}. In logistics, OCR automates data extraction from shipping labels and packaging for rapid inventory management. Retailers implement OCR for receipt scanning and product cataloging. In manufacturing, it assists with quality control and information extraction from documents. The publishing industry utilizes OCR to digitize newspapers and archives for online accessibility \cite{almutairi2024newspaper}. The automotive sector uses it for license plate recognition and vehicle registration \cite{onim2022blpnet}. Additionally, OCR aids law enforcement in forensic analysis and identity verification, besides enhancing assistive technologies for visually impaired users and facilitating real-time text translation in augmented reality (AR) and virtual reality (VR) environments \cite{nguyen2021survey}.

Considering their widespread adoption, modern intelligent systems require fast and accurate OCR models for real-time text recognition.  The core of these systems is OCR technology supported by pre-trained AI models to transform analog text into a machine-encoded format \cite{ahlawat2020improved}.
AI-based OCR models use classification techniques to recognize and extract text from images. These models have evolved significantly since their inception \cite{chaitanyaswami2023deep}. 
Early OCR systems were built on handcrafted feature extraction, where models identified textual patterns using predefined rules and statistical methods. These approaches, including template matching and optical feature detection, performed well for structured printed text but faced difficulty with font, style, and layout variations. The advent of machine learning improved the capabilities of OCR models to learn representations from labeled data. For instance, techniques such as support vector machines (SVMs) and artificial neural networks (ANNs) enhanced OCR. However, these models required extensive feature engineering and often lacked adaptability for different languages and scripts \cite{wang2021object}.  

The shift towards deep learning has revolutionized OCR and empowered models to learn hierarchical features automatically. Convolutional neural networks (CNNs) became the foundation for printed OCR with convolutional layers to extract spatial features and classify characters \cite{choudhury2021cnn}. Similarly, different types of recurrent neural networks (RNNs), particularly long short-term memory networks (LSTMs) addressed sequential dependencies in handwritten and cursive text \cite{gui2024survey}. These methods led to enhanced recognition of connected characters. Transformer-based architectures have recently dominated OCR applications. Models such as TrOCR \cite{zhang2023extending}, ViT \cite{xue2025attention}, and CharFormer \cite{shi2022charformer} have achieved state-of-the-art results using self-attention mechanisms for contextual text interpretation. They are characterized by the ability to process multilingual content, recognize handwritten scripts, and navigate complex document layouts \cite{li2023trocr}.
Despite these breakthroughs, OCR systems still fall short of human-level reading. 
While research in this area is expanding, the absence of a unified review has resulted in a lack of clarity regarding dominant OCR architectures. Moreover, several persistent challenges in OCR applications have not been explored. Many existing literature reviews related to OCR are fragmented due to focusing on individual languages \cite{balaha2021automatic}, datasets \cite{nikolaidou2022survey}, particular scenarios \cite{chen2021text} or specific model architectures \cite{saeed2019application}. 
Given the growing demand for efficient real-world OCR solutions, a systematic evaluation of current AI-based OCR models is necessary. This review bridges existing gaps by examining the evolution of OCR technologies, their multilingual capabilities, and the primary challenges. It presents a comprehensive and up-to-date synthesis of OCR-based information extraction (IE), extending beyond previous surveys that focused on narrower aspects \cite{abdullah2023systematic, fields2024survey}. By consolidating recent model developments, real-world applications, and multilingual advancements, the review provides a structured meta-analysis of the field. It critically assesses leading architectures, outlines their strengths and limitations, and proposes future research directions to advance the state of OCR systems. The key objectives of this work are as follows: 
\begin{itemize}[left=0pt, labelsep=0.5em, align=left, itemsep=1pt, parsep=1pt]
    \item A systematic review of existing OCR research to examine state-of-the-art models and supported languages.    
    \item To assess the temporal trends in OCR models from a multilingual recognition and overall accuracy perspective.  
    \item To explore the fundamental challenges encountered by existing OCR systems. 
    \item To identify the gaps in ongoing OCR research.
    \item To propose future directions for improved cross-lingual OCR, robust error correction, and better recognition of underrepresented languages.
\end{itemize}

The remainder of this paper is organized as follows: 
Section~\ref{sec: methodology} explains the review methodology. 
The results and a detailed discussion are presented in Section~\ref{sec:results and discussion}, 
and Section~\ref{sec:conclusion} concludes the paper.

\section{Review Methodology}\label{sec: methodology}
\subsection{Problem Formulation}
Advancements in AI \cite{pandey2023ai} have led to major improvements in computer vision techniques for automatically recognizing text from images using OCR \cite{baviskar2021efficient}. This progress has introduced many OCR systems, each using unique machine-learning and deep-learning models \cite{abdullah2023systematic}. Several studies focus on evaluating specific models \cite{suthar2022cnn} or testing them on individual languages \cite{sobhi2020arabic, ziyan2021china} and datasets \cite{alaei2023document,chakrabarti2024authentication}. However, no structured review explores the core OCR models.
Despite extensive research, a clear gap remains because no comprehensive study identifies the main OCR models, their strengths, and their limitations.
A broader analysis is needed to understand the barriers to accurate OCR and  to determine which models work best for different problems. This review addresses that gap by examining existing OCR models and their performance for several languages and application domains. Furthermore, it highlights underexplored areas that require additional investigation. These findings will assist researchers in making informed decisions about model selection and advance OCR technology.

\subsection{Review Scope}
The scope of this review is defined according to the PICOC (population, intervention, comparator, outcomes, context) framework, which supports a well-structured and unbiased analysis \cite{petticrew2008systematic}. The review focuses on the development and transition of OCR models from January 2015 to January 2025.
It is important to note that AI models used in OCR have been extensively studied in other fields, for example \cite{peral2024systematic}. Therefore, this review does not explain the working mechanisms or architectural designs of OCR models. Instead, it examines their application in OCR, performance in different languages and scripts, and the challenges they face. In response to these challenges, the review proposes alternative methods to mitigate the limitations. It includes peer-reviewed journal articles and conference papers published in the last ten years and excludes gray literature, extended abstracts, literature reviews, and non-English studies. The review also omits general computer vision applications (such as self-driving cars reading signboards), non-text image analysis (e.g., satellite imagery with partial textual content), and commercial software evaluations.
\subsection{Review Protocol}
\begin{figure}[ht]
    \centering  \includegraphics[width=0.9\columnwidth]{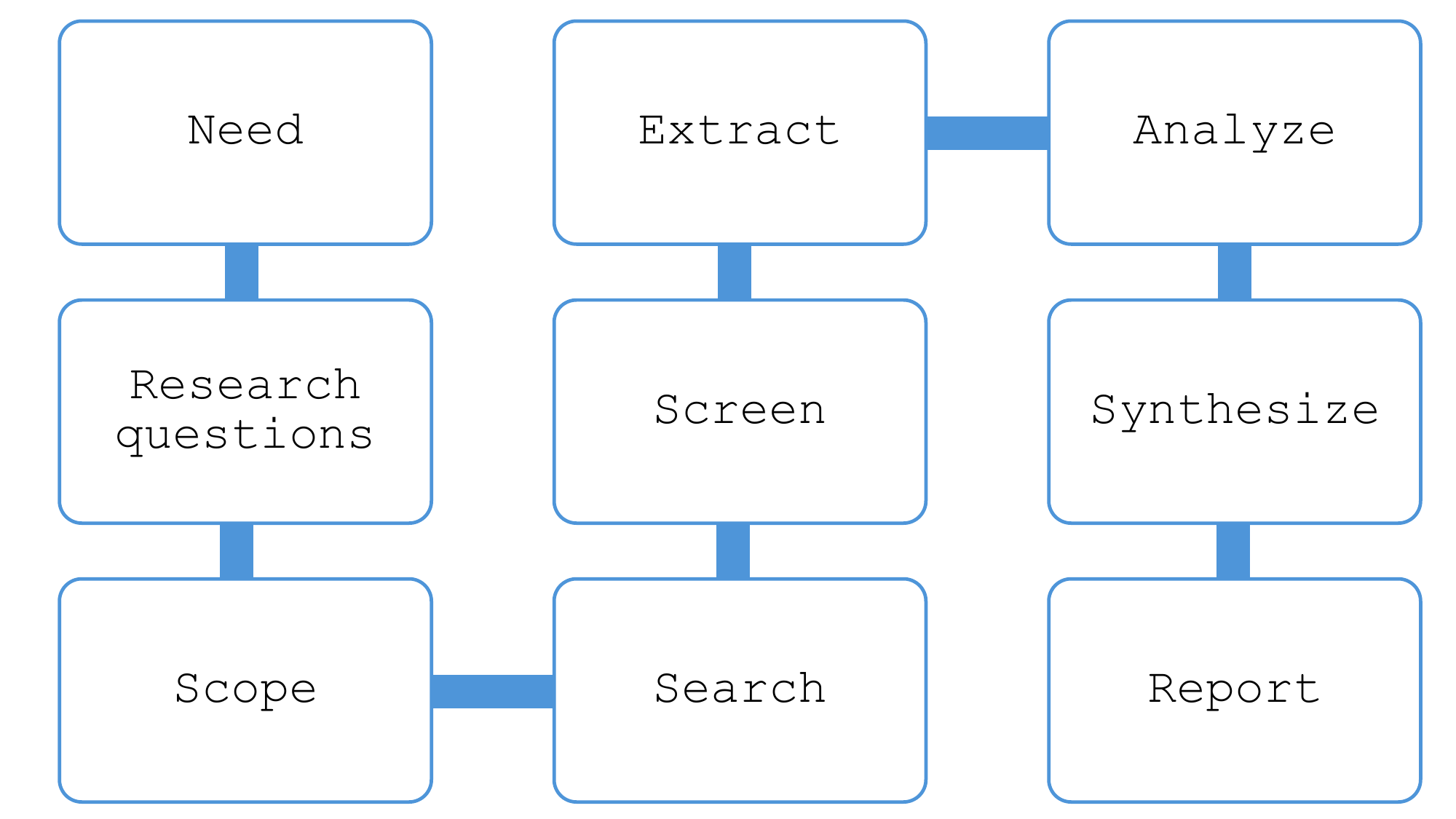}
    \caption{Review protocol.}
    \label{fig: Figure 2} 
\end{figure}

Unlike previous systematic reviews that have focused on specific aspects of OCR such as  \cite{naseer2024investigating,memon2020handwritten}, this work comprehensively covers OCR research. We conducted a systematic search across seven major academic databases:
ScienceDirect, IEEE Xplore, Scopus, Web of Science, ACM Digital Library, SpringerLink, and Google Scholar. 
These databases were selected for their broad coverage of AI, machine learning, and computer vision research. Predefined inclusion and exclusion criteria were applied to filter and select relevant studies to ensure a thorough and focused analysis.
The primary search strategy involved a combination of Boolean operators and keyword variations to maximize retrieval efficiency. The core search terms in the search string were related to OCR and AI.
\autoref{fig: Figure 2} shows the step-by-step review protocol. 

This review protocol was designed to extract, synthesize, and evaluate studies that align with the predefined research objectives. For each database, the search string was modified
to refine the search outcomes. We applied additional filters based on publication type, accessibility, and relevance to OCR model development.
By applying the inclusion and exclusion criteria, 
only papers that met the predefined selection parameters were retained for further analysis. While some papers may have indirectly addressed OCR models without explicitly mentioning key terms in their titles, abstracts, or keywords, they were omitted as they fell outside the search scope. The inclusion and exclusion criteria are outlined in \autoref{tab:criteria}.

\begin{table*}[h]
\caption{Inclusion and exclusion criteria}
    \centering
    \renewcommand{\arraystretch}{1.5}  
    \resizebox{\textwidth}{!}{
    \begin{tabular}{p{4cm} p{10cm}}
        \hline
        \textbf{Criteria Type} & \textbf{Description} \\
        \hline
        Period & Focus on recent advancements in OCR.  
        \newline Inclusion: Publications from January 2015 to January 2025.  
        \newline Exclusion: Articles published before 2015. \\
        Language & Only English language articles are included.  
        \newline Exclusion: The articles published in languages other than English. \\
        Type of literature & Only high-quality, peer-reviewed sources are considered.  
        \newline Exclusion: Grey literature such as reports, government documents, working papers, and non-peer-reviewed sources. \\
        Type of source & Only peer-reviewed journal and conference papers are included.  
        \newline Exclusion: Books, preprints, and non-peer-reviewed articles. \\
        Accessibility & Studies must be accessible from major academic databases (IEEE Xplore, Scopus, Web of Science, SpringerLink, ScienceDirect, ACM Digital Library).  
        \newline Exclusion: Articles behind paywalls without institutional access. \\
        Relevance to RQs & Articles must contribute to at least two predefined research questions.  
        \newline Exclusion: Articles that do not address at least two research questions. \\
        \hline
    \end{tabular}}
    \label{tab:criteria}
\end{table*}
Initially, a total of 45959 records were retrieved from various databases (including 1923 from Web of Science, 9039 from Scopus, 21100 from Google Scholar, 10149 from IEEE Xplore, 2480 from SpringerLink, 3121 from ScienceDirect and 689 from ACM Digital Library).  After removing irrelevant and inaccessible articles, the database was reduced to 497 articles for title screening. Following this, 310 papers met the eligibility criteria for abstract review. Upon further assessment, only 182 papers proceeded to full-text evaluation due to the inaccessibility of 128 articles. 
During this phase, we manually excluded duplicate entries and studies that lacked transparent OCR models or evaluation methodologies.
\subsection{Search Strategy and Database Selection}

\begin{figure}[t]
\includegraphics[width=1\columnwidth]{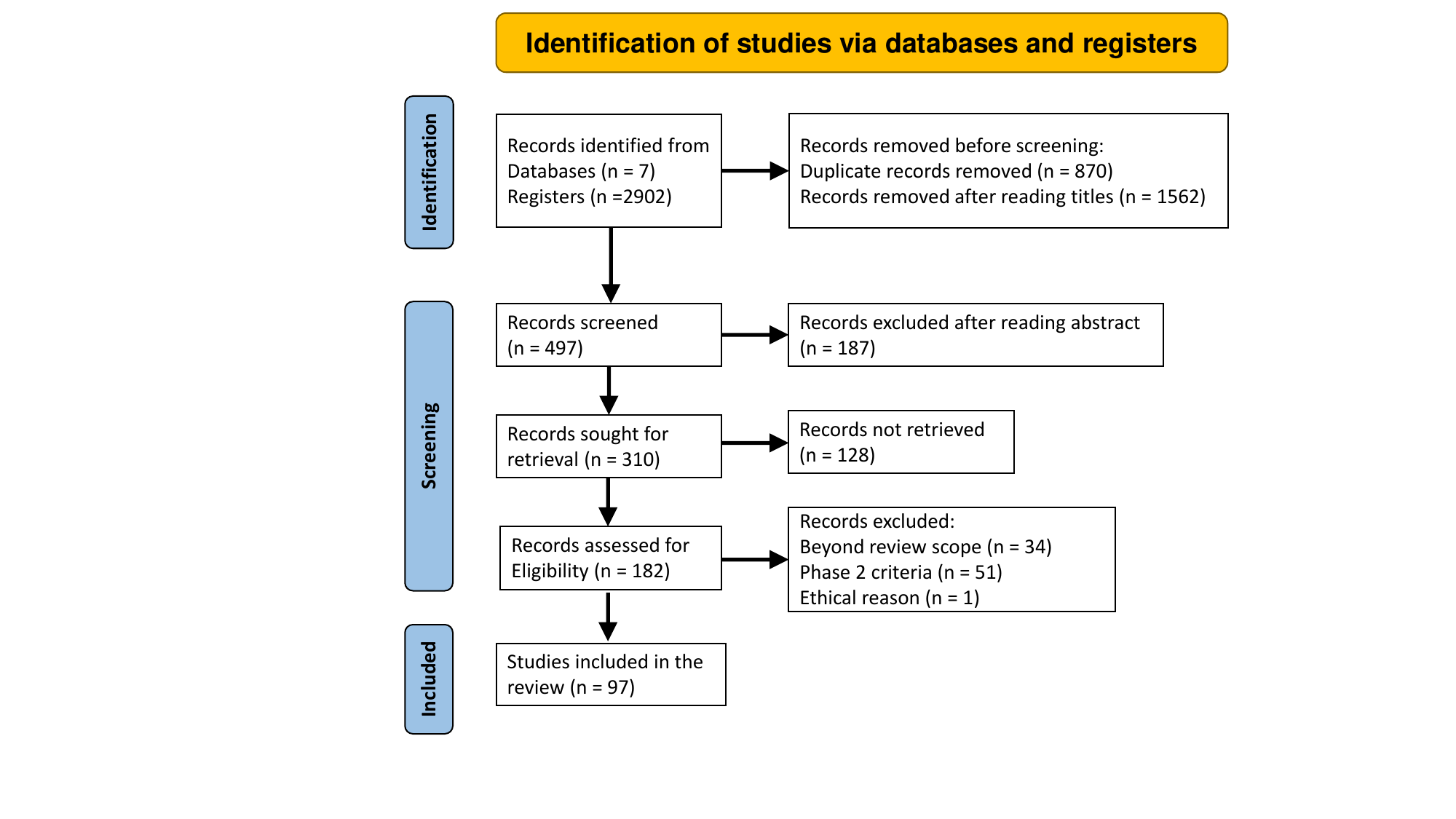}
    \caption{PRISMA flow diagram of reviewed literature.}
    \label{fig: Figure 3}
\end{figure}

\begin{figure}[t]
    \centering
    \includegraphics[width=1.2\columnwidth]{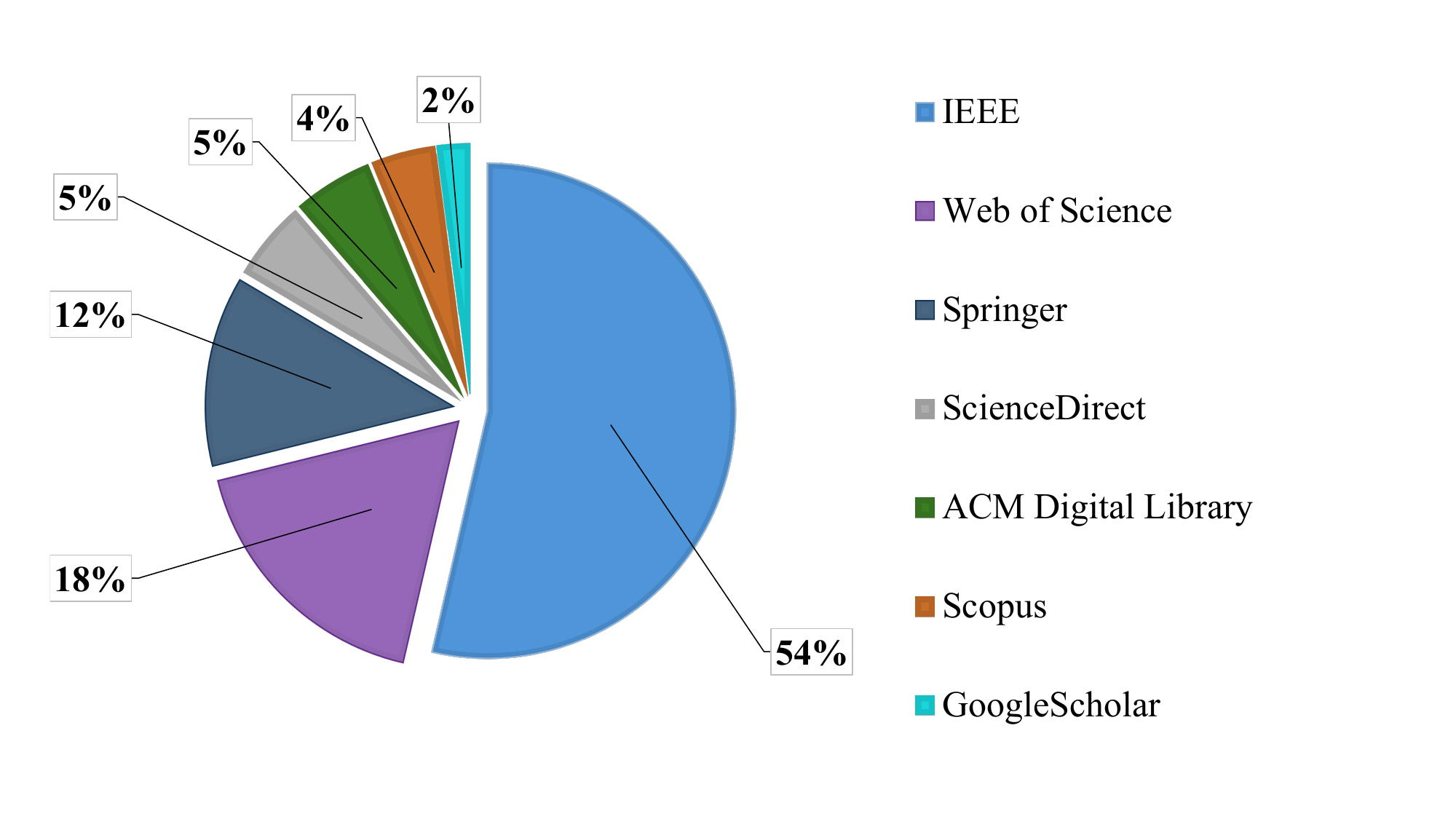}
    \caption{Database distribution of the final 97 retrieved articles after systematic screening.}
    \label{fig: Figure 4}
\end{figure}

To retrieve relevant high-quality articles, the search involved multiple iterations in refining keyword combinations. The primary search string included variations of following terms:  
("optical character recognition" OR "OCR" OR "text recognition" OR "document digitization" OR "document digitisation" OR "image text extraction" OR "character recognition") AND ("deep learning" OR "machine vision" OR "machine learning" OR Model). Searches were performed between 15–26 January 2025, applying language, date, and article type filters specific to each database. Following the inclusion and exclusion criteria, initial search results were filtered using a full-text screening process. The final dataset selection process is summarized in \autoref{tab:Table 1}. After duplicate removal and full-text analysis, the final dataset comprised 97 articles that met the predefined inclusion criteria. The PRISMA flow diagram in \autoref{fig: Figure 3} illustrates the step-by-step article selection process. Additionally, distribution of the final database of 97 article with percentages is given in \autoref{fig: Figure 4}.

\begin{table}[h]
\centering
\caption{Studies retrieved from each database in automated search}
\renewcommand{\arraystretch}{1.5}  
\resizebox{\columnwidth}{!}{  
\begin{tabular}{p{0.35\columnwidth} c c}
    \hline
    \textbf{Source} & \textbf{Initial Results} & \textbf{Retrieved Articles} \\
    \hline
    Web of Science & 1923 & 870 \\  
    Scopus & 9039 & 685 \\  
    Google Scholar & 21100 & 40 \\  
    IEEE Xplore & 10149 & 689 \\  
    SpringerLink & 2480 & 509 \\  
    ScienceDirect & 3121 & 88 \\  
    ACM Digital Library & 627 & 21 \\  
    \hline
\end{tabular}}
\label{tab:Table 1}
\end{table}

\subsection{Research Questions and Motivation}
For this systematic review, we defined three research questions (RQs) that 
align with the study’s primary objectives. These RQs are given below:  \begin{itemize}[left=0pt, labelsep=0.5em, align=left, itemsep=1pt, parsep=1pt]

    \item RQ1: What are the dominant AI-based models used in OCR, and how do they compare in terms of accuracy, efficiency, and multilingual performance?  
    \item RQ2: Which writing scripts are well-supported in OCR research, and what gaps exist?  
    \item RQ3: What are the significant challenges in current OCR systems, and what future research directions can address these issues?  
\end{itemize}

\hspace*{\parindent}Addressing these questions comes with individual and collective benefits. Individually, each question targets a critical dimension of OCR: RQ1 enhances understanding of AI-driven OCR models, RQ2 evaluates script coverage to identify inclusivity gaps, and RQ3 uncovers existing challenges to inform future advancements. Collectively, these inquiries form a comprehensive framework for an improved OCR technology. The combined outcomes will contribute to improving OCR accuracy and efficiency by addressing current limitations.

\subsection{Threats to validity}

The quality of a systematic review relies on thoroughly addressing potential threats to validity \cite{ampatzoglou2019identifying}. In this study, key threats included selection bias, information bias, and biases in the data extraction process. To minimize selection bias, we employed a broad search string followed by a selection strategy based on the method adopted in \cite{zhou2016map}. During the literature search, no keywords related to a certain model, method, or language were included in the search string. OCR research is captured equally, with statistics analyzed afterward from the final results to show a true research distribution. A second phase of selection criteria was introduced to address information bias, in which studies with ambiguous or missing information were excluded after a full-text review. Additionally, both manual and automated searches were conducted collectively on multiple databases, focusing on publication titles, abstracts, and keywords for thorough coverage of relevant studies.

While the search was restricted to English-language publications within a specific time frame, the large number of identified studies reflects the thoroughness of our approach. A data template was developed to guide information collection and reduce extraction bias. Furthermore, two researchers independently extracted the data, adhering to PRISMA guidelines to ensure consistency throughout the process and prevent disagreements.
\section{results and discussion} \label{sec:results and discussion} 
\begin{figure}[h]
    \centering
\includegraphics[width=1\linewidth]{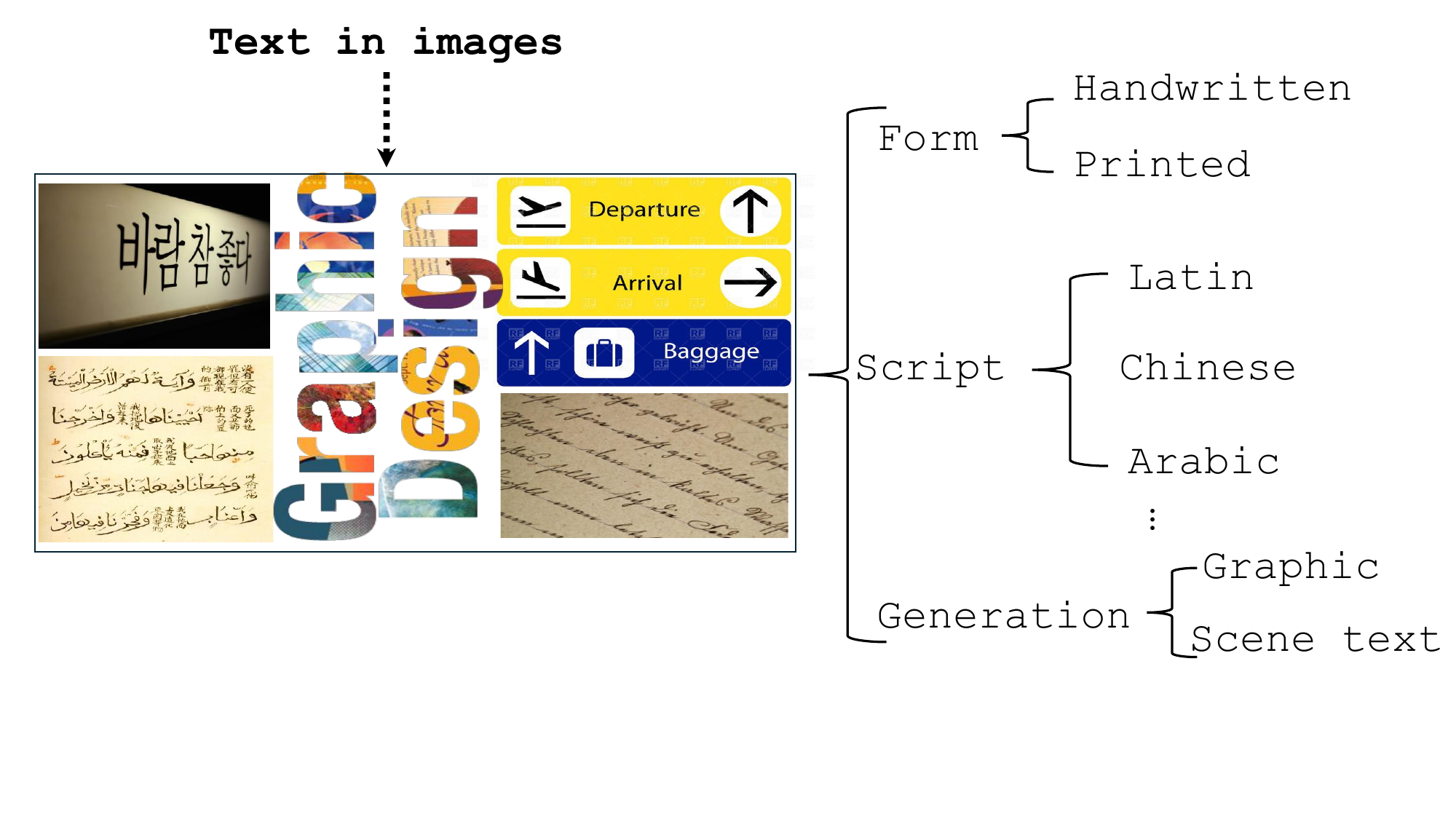}
\vspace{-4em}
\caption{Classification of text types in OCR.}
    \label{fig: Figure 5}
\end{figure}

The overview of the initially retrieved database returned prominent OCR application domains, which are broadly classified into printed and handwritten OCR. Given the multidisciplinary nature of machine vision, it is clear that machines encounter various types of text in different languages and scripts for digitization \cite{raiaan2024review}. OCR research is driven by the growing demand in fields including automatic data entry \cite{ha2022information}, license plate recognition \cite{silva2020real}, passport verification \cite{chakrabarti2024authentication}, reading traffic signs for automated systems \cite{sirohi2020convolutional}, managing traffic flow \cite{anedda2023social}, converting handwriting in pen computing \cite{alzubaidi2021real} and text to speech conversion to assist individuals with vision impairments \cite{revelli2022automate}. The research published throughout the study period initially focused on OCR model development and data preprocessing methods \cite{imam2022ocr}. From 2016 onward, the OCR applications spectrum expanded in handwritten text recognition \cite{najam2024scarce} and vehicle plate identification \cite{silva2020real}. The subsequent research in OCR focuses on complex data types and new application domains, as shown in Figure \autoref{fig: Figure 5}.
\vspace{-0.1cm}

\subsection{AI Models in OCR  (RQ1)}
\begin{figure*}[h]
    \hspace{1cm}
    \includegraphics[width=0.9\linewidth]{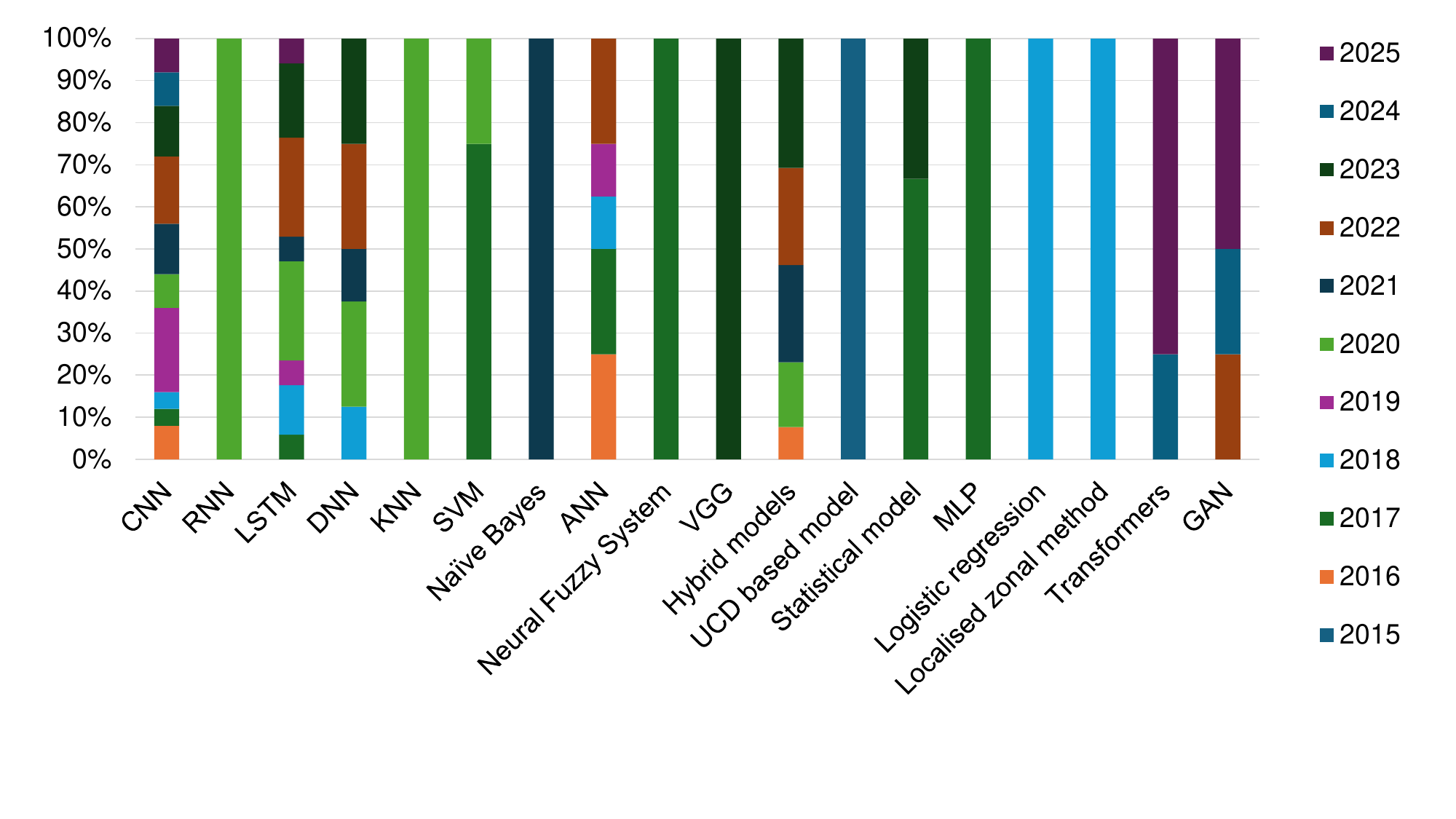}
     \vspace{-1cm}
    \caption{Prominent AI models applied to Optical Character Recognition (OCR) over the past decade.}
    \label{fig: Figure 6}
\end{figure*}

\begin{figure}[h]
    \includegraphics[width=1\linewidth]{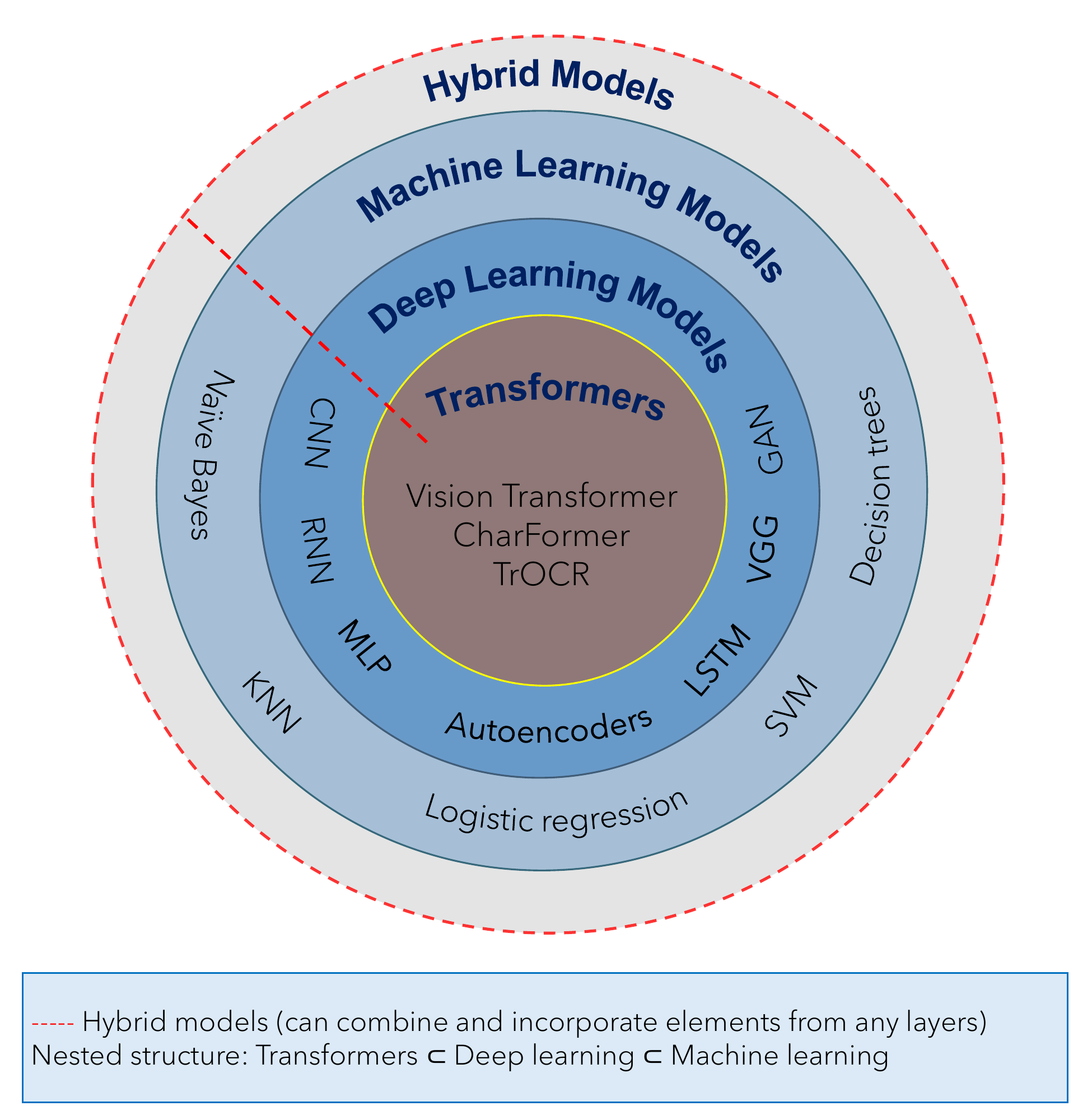}
    \caption{Evolution of OCR models architectures in a decade.}
    \label{fig: Figure 7}
\end{figure}

Classification models form the core of most OCR systems  \cite{das2020mjcn}, with text recognition typically handled using multi-class classifiers \cite{mohd2021quranic}. To analyze the evolution of OCR models over time, a manual review of the final 97 selected studies was conducted. \autoref{fig: Figure 6} presents the temporal trend analysis of individual models, while Figure \ref{fig: Figure 7} illustrates broader classification trends, grouping OCR models into four main categories: conventional machine learning models, deep learning models, transformers, and hybrid models.

 \vspace{1em}

\vspace{-0.2cm}
\subsubsection{DOMINANT AND EMERGING DEEP LEARNING MODELS (CNN, LSTM, GAN, TRANSFORMERS)}
       Since 2015, CNNs have been widely used in image-based recognition tasks \cite{suthar2022cnn}. Different CNN architectures experienced steady growth and remained dominant until a surge of handwritten text. LSTM networks gained prominence after 2017 \cite{zou2020robust} for sequential and bidirectional text. However, their popularity has slightly declined in recent years, likely due to the rise of more advanced alternatives. Since 2020, GANs have attracted increasing attention in OCR research primarily for their ability to generate realistic synthetic data, enhance low-quality images, and refine recognition outputs. Although GANs do not directly perform OCR classification, they are valuable auxiliary techniques to support other OCR models \cite{agrawal2024exploration}. More recently, OCR has been dominated by Transformer-based deep learning architectures. Since 2020, Transformers have seen significant growth and have largely replaced LSTMs in many OCR-related applications \cite{azadbakht2022multipath}. With dynamic feature selection, Transformers are a superior alternative to LSTMs for efficiently recognizing sequential data like handwritten text. 
    
\subsubsection{ADAPTIVE AND EVOLVING HYBRID MODELS AND NEURAL FUZZY SYSTEMS}
Hybrid OCR models are important when standalone models deliver suboptimal results. These models enhance text recognition by combining different approaches, especially in cases where a single method is inadequate due to limited training data or interpretation challenges. When dealing with distorted or ambiguous text, rule-based techniques refine predictions. A great example is Neural Fuzzy Systems, which have gained attention since 2020 for their ability to handle uncertainty and improve decision-making in complex recognition tasks. By integrating fuzzy logic with other learning models, these models efficiently process noisy or highly variable text \cite{sahlol2020handwritten}. Hybrid models are prioritized for recognizing text in languages and scripts with limited training data where characters can appear in different forms.
    
\subsubsection{DECLINING TRADITIONAL MACHINE LEARNING MODELS}
Before 2019, machine learning models like KNN, SVM, Naïve Bayes, Logistic regression, and statistical models were widely used in OCR \cite{abhishek2020optical, raj2015optical}. Their appeal came from their ability to recognize text even with limited training data. These models were ideal for early OCR systems when computational resources were scarce and large labeled datasets were hard to obtain. KNN and SVM are commonly applied to character recognition due to their strong classification capabilities, while Naïve Bayes is practical for text-based OCR. At that time, ANN and MLP, as early neural network-based models, helped with pattern recognition \cite{abhishek2020optical}. At the same time, logistic regression and statistical models were also used in rule-based approaches to text segmentation and classification. These models excelled in OCR on structured, small-scale datasets where handcrafted features were engineered for specific tasks \cite{mahastama2020optical}.

However, as OCR tasks grew more complex and the availability of big data increased, the limitations of traditional machine learning became evident. These models require extensive manual feature engineering, which demands domain expertise. While they worked well on simple datasets, they encountered obstacles with large-scale, high-dimensional data, when variations in fonts, handwriting styles, and document layouts introduced complexity beyond their capabilities. As a result, the reliance on traditional machine learning models has steadily declined.

\subsection{TRANSITION IN OCR APPLICATIONS SPECTRUM}
\begin{figure}[h]
    \centering
   \hspace{0.1cm}   
   \includegraphics[width=1.05\linewidth]{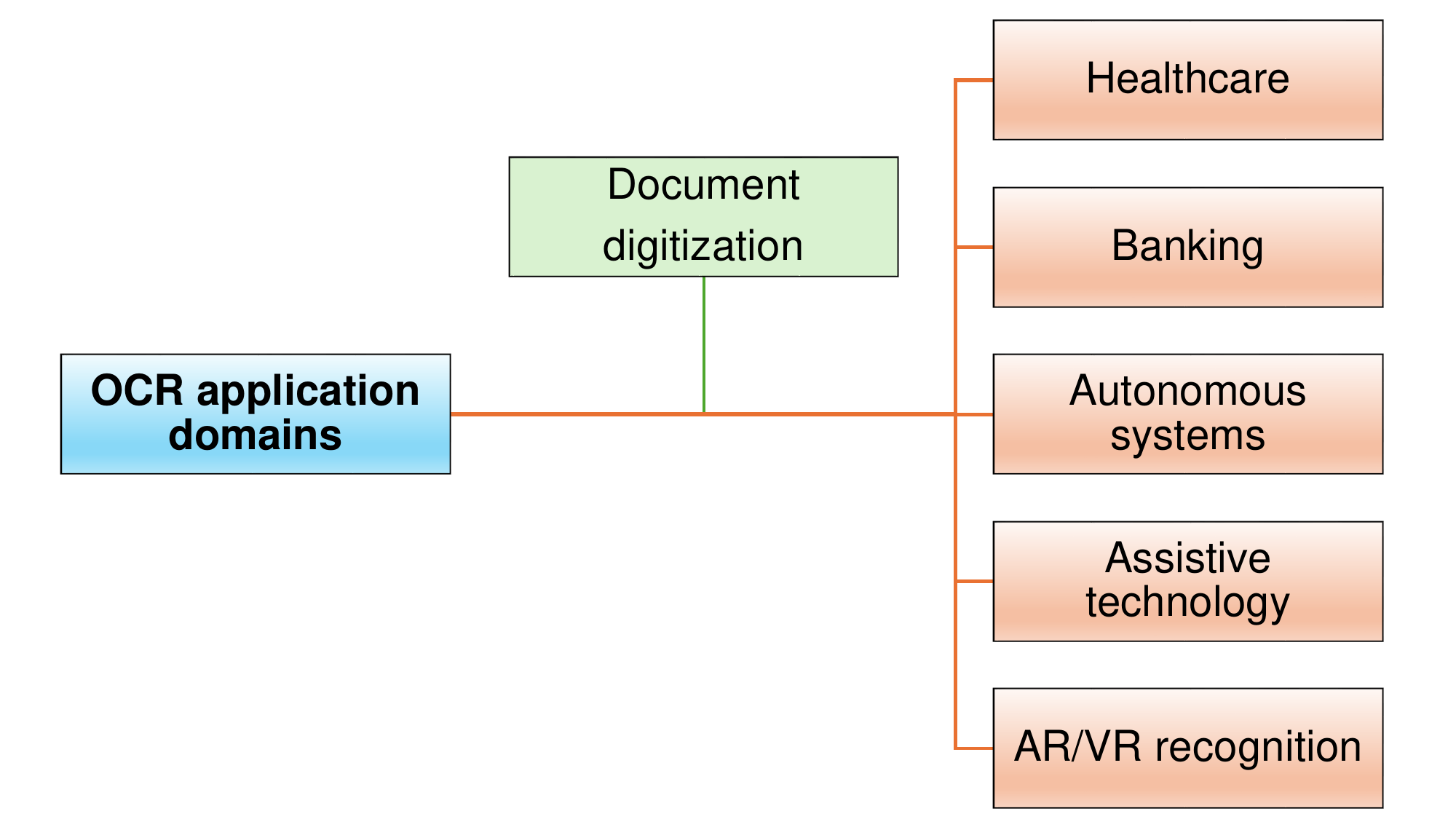}
    \caption{Popular OCR application domains. General domains (blue box) initially focused on document digitization (green box) and expanded to other domains (orange boxes).}

    \label{fig: Figure 8}
\end{figure}

The applications of OCR reflect the extensive research and growing popularity of modern OCR systems. Recent research has oriented towards integrating OCR with advanced AI-driven applications shown in \autoref{fig: Figure 8}. OCR, which started with basic document digitization has moved toward a deeper understanding of documents in healthcare and banking \cite{francis2025comparison}. Now, autonomous systems extract text from real-world scenes with higher accuracy. 
On the other hand, multilingual recognition improves translation for low-resource languages \cite{ghafoor2021impact}. Assistive technologies use OCR to support visually impaired individuals in reading and navigation \cite{arakeri2018assistive}. Recently, multi-modal learning in augmented reality (AR)/virtual reality (VR) has been used for user engagement in an immersive environment \cite{cao2023mobile}. Advanced research beyond applied models now addresses complex challenges, broader applications, and the need for stronger, more adaptable systems \cite{mahadevkar2024exploring}. 
Recent publications have mainly focused on multilingual, multi-modal, and high-performance OCR solutions \cite{sanfilippo2021multi}.

The AI models in OCR research have evolved from traditional statistical methods to deep learning techniques for higher performance. The performance of OCR is typically evaluated using a variety of metrics including accuracy, precision, recall, F1-score, speed, and specialized metrics like character error rate (CER) and word error rate (WER). 
However, due to the variations in text formats ranging from individual digits and characters to complete words and sentences, accuracy remains the most consistently reported and broadly applicable evaluation metric in OCR research.  Table \ref{tab:combined_ocr_models} presents the reported accuracy values of each model on different datasets. The minimum and maximum accuracy values reflect the performance range observed across separate studies using the same model architecture. These results were extracted directly from the original papers, without averaging or statistical aggregation. It is important to note that studies use various types and amounts of data for training and testing their methods, pre-process and split datasets in different ways. Therefore, it is impractical to suggest the best performing model based on direct comparison. As such, the table describes the reported performance of each model. Early models like logistic regression and KNN faced complex scripts and handwriting challenges \cite{arora2024devanagari}. The low accuracy of conventional machine learning led to the adoption of CNNs and LSTMs for OCR tasks. Deep learning significantly improved feature extraction and sequence modeling. Hybrid architectures (CNN-LSTM, CNN-RNN, CNN-YOLO) have been explored to combine feature extraction and sequence learning capabilities. It resulted in higher accuracy than standalone models \cite{hangaragi2023accuracy}. However, complexity often increases training time and computational costs. Hence, it makes them less feasible for large-scale OCR applications. 
Hybrid models combining statistical and deep learning techniques are proposed in noisy document scenarios \cite{huang2021separating}. Despite improvements, challenges persist for low-resource languages and complex scripts because of the limited availability of training data.

In 2019, CNNs transformed OCR with high accuracy, while hybrid models like RF-CNN improved flexibility. By 2020, CNNs dominated printed text recognition, with LSTMs and CNN-LSTM hybrids excelling in sequential data recognition. Lightweight machine learning models still play an important role in resource-constrained applications.
Initially developed for synthetic data generation, GANs found success for OCR in low-resource or degraded text scenarios. Their exceptional data generation feature is beneficial in OCR when handling incomplete, noisy, or distorted text data \cite{ibrahim2022license}. 

Deep learning-based transformer models have recently gained prominence for their superior contextual understanding and multilingual capabilities despite their high computational requirements \cite{xu2023multimodal}. 
Transformers use attention mechanisms to understand character and word relationships. These models, such as Vision Transformers (ViT) and TrOCR, have already improved efficiency through direct image-to-text conversion. AI models using these mechanisms have achieved an impressive 99.89\%  accuracy in character recognition\cite{kumar2023attention}.

The main difference between conventional machine learning, conventional deep learning, and transformers is how they handle data, feature extraction, context, and scalability. Traditional machine learning models rely on handcrafted features and do not inherently capture sequential dependencies \cite{azam2023comparative}. They perform well on structured data but underperform with complex patterns in large-scale unstructured data. Conventional deep learning models learn features automatically but have limitations in processing long-range dependencies. RNN-based deep learning models like LSTMs handle sequential data but suffer from vanishing gradients, sequential bottlenecks, and inefficient scalability \cite{ahmed2023deep}. 

Transformers, in contrast, overcome these challenges with self-attention,  parallel processing, and better long-range dependency capture \cite{gu2024transliterated}. Unlike conventional deep learning, transformers do not rely on recurrence. Hence, they are faster for OCR tasks that require contextual understanding. However, this comes at the cost of higher memory consumption and the need for large datasets for optimal performance \cite{mortadi2023alnasikh}.

\begin{table*}[h]
\centering
\caption{Reported accuracy of each OCR model on various datasets from selected studies}
\renewcommand{\arraystretch}{1.5} 
\resizebox{\textwidth}{!}{ 
\begin{tabular}{p{2.2cm} p{3cm} p{3cm} p{2.5cm} p{2.2cm} p{3.5cm}}
    \hline
    \textbf{Model Type} & \textbf{Architecture} & \textbf{Datasets Used} & \textbf{Text Type / Language(s)}  & \textbf{Performance (Accuracy)} &\textbf{Strength / Limitation} \\
    \hline
    SVM \cite{yamina2017printed},\cite{abhishek2020optical},\cite{das2021ensemble},\cite{phangtriastu2017comparison}, & SVM-MLP-ETC, NCC, HOG-SVM & 286 Font Packs, Bangla Handwritten Dataset & Printed / Handwritten (English, Bangla) & 60.6\% - 94.53\% & Works well for structured fonts but struggles with handwritten text and complex scripts \\  
    ANN \cite{jing2017efficient},\cite{ afroge2016optical},\cite{hussien2015optical, lee2018optical} & Feedforward NN, Backpropagation NN, Hopfield-ANN & MNIST, CROHME, EMNIST, Picture Editor Dataset & Printed / Handwritten (English, Arabic)  &   92\% - 100\% &  Effective in basic OCR but lacks generalization for large-scale real-world datasets \\  
    Neural Fuzzy System \cite{bhyrapuneni2022comparative, maung2017myanmar} & IT2NFS-SPSA, ANFIS & Alphabetic and Numeric Datasets, Burmese OCR & Printed (English, Burmese)  & 7.25\% - 98.20\% & Reduces computational cost compared to traditional fuzzy systems but limited in scalability \\  
    CNN \cite{busa2022small, shukla2023devnagari, zaryab2023optical, mohd2021quranic, suthar2022cnn, dessai2019deep} & VGG16, CRNN, Inception V3, ResNet50, Multi-CNN & MJSynth, Kaggle, IAM, BanglaLekha-Isolated, CMATERdb, EMNIST & Printed/Handwritten (English, Arabic, Tamil, Bangla, Gujarati, Devanagari) &57.2\% - 99.76\%  &  Strong for printed OCR, but struggles with handwritten cursive text and script complexity \\  
LSTM/BLSTM \cite{zou2020robust},\cite{onim2022blpnet}, \cite{singh2019real},  \cite{serhan2023gpu},\cite{wasan2022development}, \cite{gener2022gpgpu}, \cite{alamsyah2022autoencoder}, \cite{scazzoli2019usage}, \cite{akinbade2020adaptive}, \cite{yanfi2024multi} & Bi-LSTM, Gated-CNN-BLSTM, Tesseract LSTM & Medical Records, EMNIST, IAM, BCSD & Long-Sequenced Text (English, Arabic, German, Indonesian) & 91\% - 99.94\% & Worked well for long-sequence text, but require high computational resources \\  
    Transformer\cite{rakshit2023novel, azadbakht2022multipath, zaryab2023optical, shi2022charformer}  & Vision Transformer (ViT), TrOCR, CharFormer & Shotor, Sadri, Born-Digital, LicenseNet & Multilingual (Chinese, English, Persian) & 77.25\% - 99.98\%  & Outperforms CNNs/LSTMs in multilingual OCR, but computationally expensive \\  
    GAN \cite{ibrahim2022license, huang2021separating} & CycleGAN, SRGAN, DESRGAN & CASIA HWDB, Synthetic Data & Noise-Resistant OCR (Chinese, English) & 87.4\% - 99.89\%  & Effective for enhancing resolution of noisy images but prioritizes realism over textual correctness. \\  
    Hybrid Models \cite{choudhury2021cnn,dhanikonda2022efficient,zhao2020improving} & CNN-LSTM, CNN-RNN, CNN-YOLO, Faster R-CNN & IIIT-HW-Telugu, CMATERdb, LicenseNet, CCPD & Printed/Handwritten (English, Arabic, Telugu, Chinese)& 94.5\% - 99.75\% & Combining different models improves accuracy. However, training complexity increases \\  
    \hline
\end{tabular}}
\label{tab:combined_ocr_models}
\end{table*}
\subsection{PERFORMANCE OF OCR MODELS ON HETEROGENEOUS DATASETS}
When the data is complicated,  OCR models evolved from rule-based and statistical models to deep learning architectures. As the architectures of AI models contrast, their accuracy on various datasets differs; therefore, instead of a direct comparison, a comprehensive assessment of all models is tabulated in \autoref{tab:combined_ocr_models} for better understanding. Accuracy was chosen as the primary evaluation metric due to its consistent reporting in 72\% of studies. In contrast, key OCR metrics such as WER and CER appeared in only 5\% and 4\% of studies, respectively. The limited reporting of other metrics prevented their use in cross-study analysis. In the same way, presenting an in-depth analysis of popular datasets and data processing methods merits writing another comprehensive literature review.  From current analysis, it is found that OCR models are capable of recognizing printed and handwritten text containing symbols, digits, and characters in many languages. The performance gap between traditional and deep learning-based OCR models is evident from the accuracy range of each model. In SVM model a significant performance significant performance variation was observed. The accuracy ranged from 60\% to 94\% depending on script complexity, model architecture, hyperparameters, and dataset characteristics. Overall variation in accuracy indicates SVM can handle simpler OCR tasks, but its performance degrades for complex scripts, handwritten text, or degraded image quality. Similarly, CNN-based models like VGG16 and CRNN demonstrate consistent performance on printed text datasets but show notable accuracy drops on handwritten text due to their inherent limitations in processing sequential information and character dependencies. Compared to SVM and ANN-based models, deep learning-based transformer models, Vision Transformer (ViT), TrOCR, and CharForme maintain higher accuracy above 77\% regardless of the datasets. 

More recent Transformer-based OCR models outperform CNNs and LSTMs for multilingual and complex-script OCR due to their self-attention mechanisms. Transformer-based models achieved high accuracy and demonstrated state-of-the-art performance for languages with complex diacritics. Multi-head attention, parallel processing, bidirectional representation and self-supervised learning are distinguishing characteristics of Transformers \cite{khan2022transformers}. However, these models are computationally expensive as compared to traditional CNN architectures. For instance, TrOCR \cite{li2023trocr} uses pre-trained Vision Transformers for both image encoding and autoregressive decoding. The base and large versions of this model contain 334 million and 558 million parameters, respectively. These sizes are up to 20 times larger than those of traditional CNN-based models, which typically contain between 10 million and 50 million parameters. As a result, memory demand of TrOCR is high with usage ranging from 4 to 16 gigabytes, compared to the 500 megabytes to 2 gigabytes required by CNN-based models. This also leads to longer inference times, which can range from 100 to 500 milliseconds, whereas CNNs typically perform inference within 10 to 50 milliseconds. Additionally, the training costs are considerably higher for TrOCR. Similarly, Donut \cite{kim2022ocr} avoids traditional OCR pipelines by relying on Transformer-based architectures. However, it requires over 200 million parameters and encounters similar computational and memory challenges. The self-attention mechanism with quadratic memory complexity necessitates the use of GPUs with at least 16 gigabytes of VRAM. It also requires a substantially higher number of floating-point operations per image, ranging from 10 to over 100 gigaflops, compared to the 1 to 10 gigaflops typically required by CNNs. Therefore, the feasibility of transformers in real-time and mobile applications is limited. For an in-depth comparison between Transformers and CNNs, along with a detailed exposition of the technical foundations of Transformer architectures, refer to study \cite{khan2022transformers}.  

\subsubsection{CORRELATION BETWEEN INPUT DATA AND GENERALIZATION OF OCR MODELS}
The overall OCR accuracy is driven by several fundamental, architectural, and universal factors.
 Apart from the architecture of the classification model, hybridization, hyperparameter tuning, data preprocessing steps, training strategies, and post-processing methods, the ultimate performance of an OCR system is determined by the dataset complexities. Existing OCR models underperform in complex dynamic environments due to the complexity of the background information of the input data \cite{tzoumpas2024data}. Pretraining data coverage is another major factor due to the limited transferability of pretrained models. Models trained primarily on Latin scripts may underperform on Arabic, Chinese, or Devanagari scripts due to differences in text direction (left-to-right vs. right-to-left), character complexity (e.g., dots, diacritics, and strokes), and word structure (spaced vs. compound). The reverse is also true for models trained on non-Latin scripts. Models trained on varied datasets, including multilingual, handwritten, or historical documents, tend to generalize better in real-world scenarios.

Real scene images contain text along with background, foreground information, and noise. It is challenging to detect, segment, and extract text from low-contrast real scene images with partial visibility. In addition to degraded images and noisy backgrounds, real-world OCR tasks frequently involve more complex visual conditions such as skewed perspectives, motion blur, occlusion, and multilingual text appearing within the same scene \cite{xu2023multimodal}. These adversarial conditions challenge OCR models when the training data does not reflect such variability. Several real-scene datasets like ICDAR robust reading challenge, ICDAR incidental scene text, Street View Text (SVT), COCO-Text, SynthText, MJSynth, and TextOCR have been developed to capture the complexities of multilingual scripts, varied lighting conditions, handwriting variability and natural image distortions. However, a major limitation of these datasets is the absence of associated lexicons, which restricts the development and evaluation of OCR models that rely on language constraints or post-processing for improved accuracy \cite{chen2021text}. The scale and script diversity in these datasets is limited compared to the complexity in real unconstrained environments. In such contexts, OCR accuracy is enhanced by combining domain-specific data augmentation, advanced feature learning, and semi-supervised techniques to bridge the gap between controlled training data and uncontrolled real-world inputs \cite{gbada2025deep}. 
\subsubsection{CONTEXTUAL SENSITIVITY OF OCR MODELS}
The accuracy variations of the same model in different studies indicate that OCR performance depends on architecture, training strategy, data coverage, and decoding method. This includes how well a model's architecture aligns with the training design \cite{martinek2020building}. Earlier OCR systems, based on machine learning techniques, lacked the flexibility needed for complex or noisy layouts. While Transformer-based models show consistent advantages over traditional machine learning methods and CNN-based architecture for complex images \cite{anand2025winograd}. The ability to model contextual relationships and long-range dependencies between characters leads to higher recognition accuracy on multi-column and irregularly formatted documents \cite{li2023trocr}. Transformer-based models are suitable for visually complex or noisy inputs, while CNNs are preferred in resource-limited environments with high-quality inputs \cite{aliyu2024sentiment}. Hybrid systems, which combine pretraining, flexible decoding, and language-aware components, are well-suited for multilingual or high-accuracy applications such as text script identification \cite{yang2025script}. Data augmentation with geometric transformations like rotation and perspective changes improves generalization on distorted inputs. Besides geometric transformations, augmentation strategies such as color space transformations, kernel filters, mixing images, random erasing, feature space augmentation, adversarial training, GAN-based augmentation, neural style transfer, and meta-learning schemes are good solutions for positional biases present in the training data \cite{shorten2019survey}. Research suggests that different models respond differently to different preprocessing steps \cite{amiriebrahimabadi2024comprehensive}. Traditional machine learning approaches are comparatively more reliant on binarization, denoising, skew correction, feature extraction, and segmentation. Preprocessing reduces input complexity and enhances feature clarity. Classical models require clean, well-aligned inputs and handcrafted features. The performance of these models is strongly influenced by preprocessing quality, and their generalization is limited when inputs are noisy or documents deviate from the expected layout. In addition, hyperparameters such as kernel functions in SVM, the number of neighbors in k-NN, and decision criteria in tree-based models are manually selected or externally tuned using techniques like grid search or cross-validation. These models are also sensitive to overfitting when trained on very large, very small, or imbalanced datasets \cite{sharifani2023machine}. In contrast, deep learning models learn hierarchical feature representations directly from raw input pixels. Their dependency on manual feature engineering is limited. However, their performance relies on the availability of large, well-labeled datasets \cite{gbada2025deep}. Without sufficient data, deep learning models are prone to overfitting when trained from scratch. In such cases, data augmentation becomes an important strategy with techniques such as random rotation, scaling, contrast shifts, and synthetic noise, etc, to simulate variability found in real-world documents. Performance of deep learning models varies with changing training configurations. Hyperparameters such as learning rate, optimizer type, batch size, network depth, and regularization strategies, for instance, dropout, batch normalization, weight decay, etc., influence OCR accuracy \cite{choudhury2021cnn}. To improve convergence and generalization in deep learning models, different techniques such as early stopping, learning rate scheduling, and data normalization are used. Hybrid models are generally more robust than a model alone, especially in recognizing text lines with varying lengths and orientations. However, additional complexities are introduced during training, including hyperparameter choices \cite{liao2022empirical} and sequence length management. Stabilizing training may require techniques such as gradient clipping and careful weight initialization. In contrast, the end-to-end architectures in transformers can model entire sequences or documents using self-attention mechanisms. These models learn long-range dependencies and attend to relevant regions without explicit segmentation or handcrafted features. They perform well even on inputs with irregular layouts, noise, or partial occlusions. However, they are highly dependent on large-scale pretraining, typically on multimodal datasets, and fine-tuning on task-specific data. Factors such as input resolution, tokenization strategy, positional encoding, and optimizer configurations have a substantial influence on Transformers \cite{li2023trocr}. The relative impact of different factors on OCR varies, depending on the model family. Understanding these distinctions is critical when designing OCR systems for specific tasks or document domains.
\subsection{Text Scripts in OCR Research (RQ2)}
\begin{figure*}[h]
    \centering
    
\includegraphics[width=0.9\textwidth]{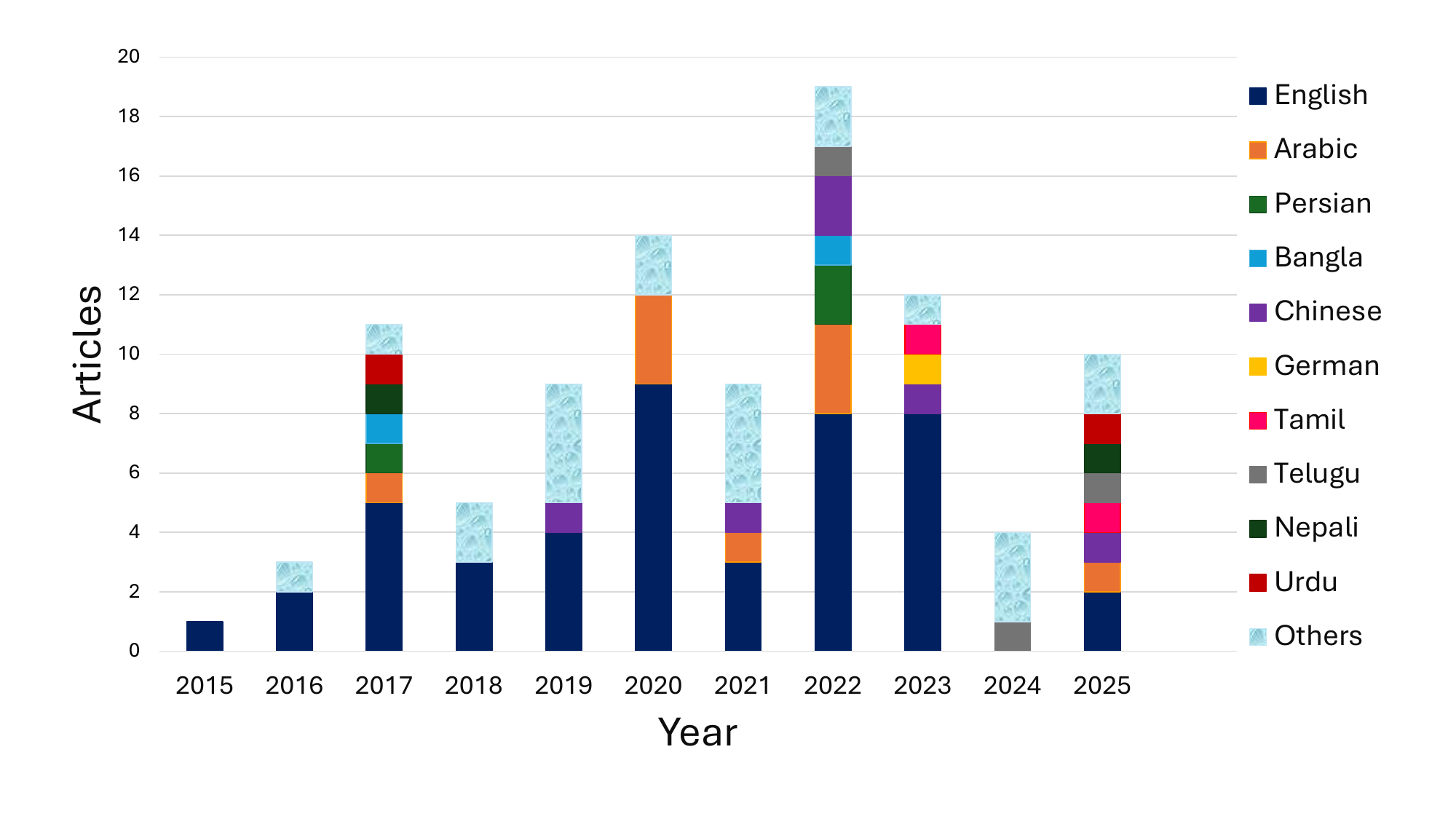}
    \vspace{-0.5cm}
    \caption{Trends in language-specific OCR models.}
    \label{fig: Figure 9}
\end{figure*}

Language is a complex vocabulary, grammar, and phonetics system while its written counterpart, script is a structured visual representation. Scripts consist of symbols, letters, or characters that transcribe spoken language into a readable format. Interestingly, a single language can be represented in multiple scripts, just as a single script can be adapted for different languages. Based on their approach to encoding linguistic units, scripts are generally classified into three main categories: alphabetic, syllabic, and logographic \cite{kuester2023learning}. 
Multilingual OCR research still has a long way to go, especially compared to how well English is doing. Since English is resource-rich and works smoothly with most existing OCR tools, it has been the most used and researched language. English is one of the most spoken languages in the world. It is the official language in 53 countries and the first language for about 400 million people. Many bilingual speakers also use it to communicate globally. Over the years, researchers have studied OCR for English more than for any other language, leading to many published studies \cite{memon2020handwritten}.
Because of this extensive research, English OCR systems have greatly improved and are widely used. Today, English OCR is mature, and its accuracy is high due to extensive training data, simple character sets, consistent spacing, and limited diacritics. After English, Arabic is the second most researched language in OCR. Written right to left, Arabic features letters that change shape based on position and diacritics that alter meaning. Complex Arabic script comprising connected characters is challenging for OCR. Similar-looking letters and variations in handwriting further complicate text recognition. These factors demand specialized OCR models for accurate processing, and research is being carried out, too.
Although the demand for multilingual OCR continues to rise, languages with complex scripts have not advanced at the same pace as those with more straightforward writing systems. Primarily, it is due to the scarcity of high-quality datasets and the inherent complexity of the scripts themselves. The less explored languages, such as Bangla, Tamil, Telugu, Oriya, Devanagari, Chinese, Gujarati, Korean, Finnish, Indonesian, Javanese, Uighur, Myanmar, and Nepali and more lack quality datasets and advanced OCR solutions.
As shown in \autoref{fig: Figure 9}, these languages lag behind English due to limited resources and lower recognition accuracy. Moreover, their harder-to-process scripts result in less accurate OCR. There is an apparent necessity to bridge the gap for the global adoption of OCR by achieving high accuracy in languages other than English. Further research is needed to develop more inclusive multilingual OCR models with enhanced recognition capabilities for underrepresented languages. 

\begin{figure*}[h]
    \centering
\includegraphics[width=0.8\textwidth]{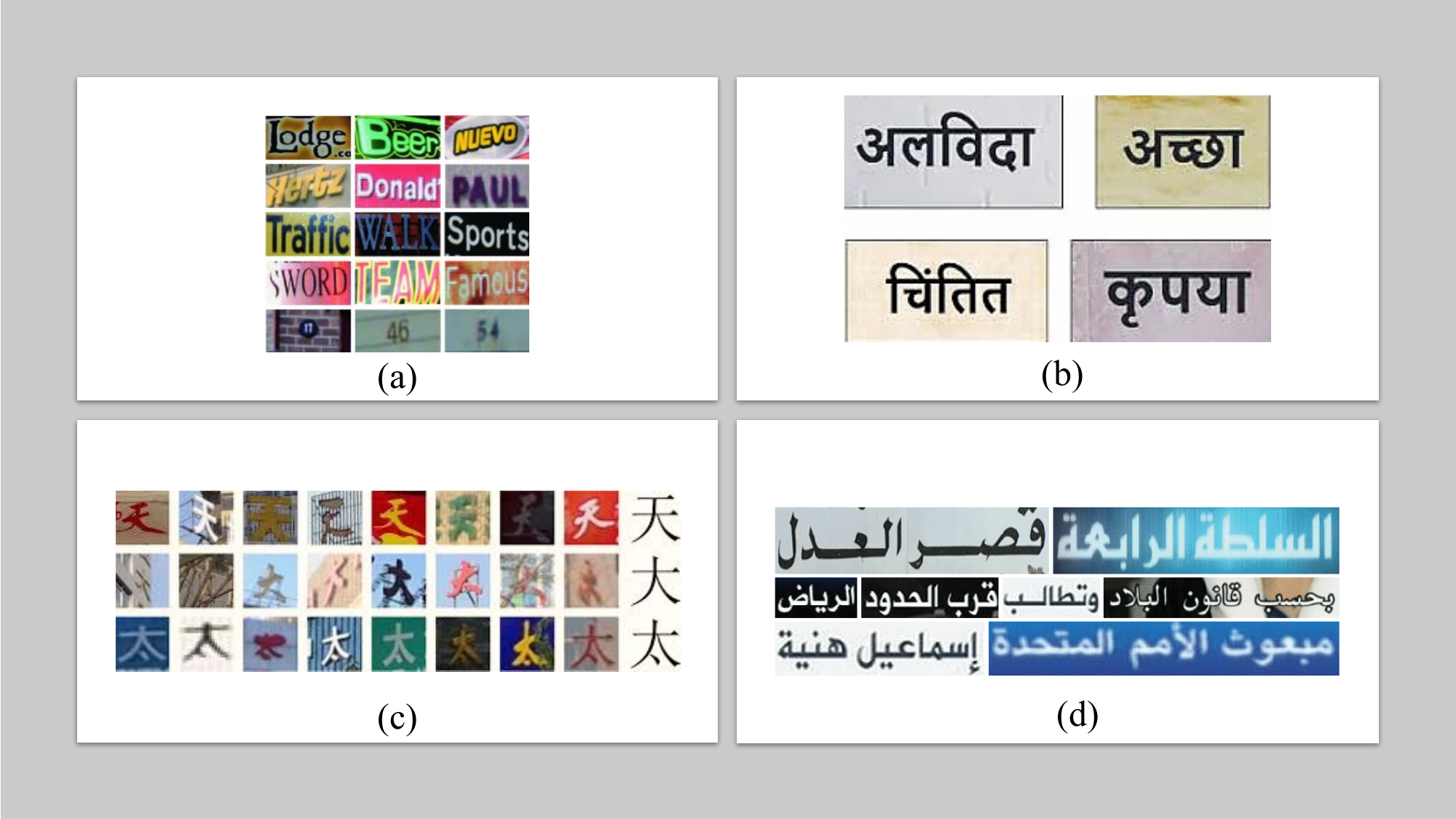}
    \caption{Sample of printed scripts: (a) Latin, (b) Devanagari, (c) Chinese, and  (d) Arabic.}
    \label{fig:Figure 10}
\end{figure*}

Current research in OCR focuses on improving multilingual script recognition. However, a one-size-fits-all OCR model is ineffective due to the vast differences in writing systems. Current script-specific OCR systems are designed for the unique structural, linguistic, and typographical characteristics of individual scripts. This review explored prominent scripts adopted in OCR research, their unique characteristics, and the languages they support. Representative samples of the most commonly reported scripts are illustrated in \autoref{fig:Figure 10} followed by a brief discussion on prominent characteristics of each script. The identified features in different scripts sharing some common characteristics can be used in grouping similar script patterns to apply the same recognition approaches. This will improve the generalization performance of OCR models.  
\subsubsection{LATIN SCRIPT}
The Latin script, also known as the Roman alphabet, has its roots in the old Italic scripts, which were influenced by the Greek alphabet and ultimately derived from the Phoenician script \cite{burlacu2021digitising}. The Romans standardized it, and it has since evolved into the world’s most widely used writing system \cite{asselborn2021transferability}. It is the foundation for hundreds of languages, including English, Spanish, French, German, Italian, Portuguese, Dutch, Turkish, Vietnamese, and many more \cite{byambadorj2021normalization}. It is the dominant writing system in Europe, the Americas, parts of Africa, and certain regions of Asia and Oceania \cite{lebsanft2020manual}.
It is one of the most versatile and enduring writing systems in history. Over time, the Latin script has evolved into numerous typographic styles, from simple print fonts to artistic calligraphy.
It consists of a finite set of characters (letters, numbers, and punctuation) with consistent shapes. Some of the defining features of the Latin script include its left-to-right writing direction, the presence of both uppercase (capital) and lowercase letters, and the explicit representation of vowels as essential characters. Latin characters have only two forms, uppercase and lowercase, that maintain shape within a word \cite{gu2024transliterated}. 
Due to these characteristics, the Latin script is highly compatible with digital processing. Individually spaced letters are easy to segment and recognize for OCR models. The Latin script has also been widely standardized through Unicode encoding for seamless digital representation.
  
\subsubsection{ARABIC SCRIPT}
The Arabic alphabetic script, known as the Arabic abjad, originates from the Nabataean script, which evolved from Aramaic around the 4th century CE \cite{gruendler2019development}. Over time, it developed into a highly cursive and adaptable writing system. It is used for many languages, either as their primary writing system or as an alternative, including Arabic, Persian, Urdu, Pashto, Kurdish, Sindhi, Punjabi, Malay (Jawi), Uyghur, and Hausa. Various languages in the Middle East, North Africa, South Asia, Central Asia, Southeast Asia, and parts of Africa are expressed in the Arabic script. Some exceptional characteristics of Arabic script are right-to-left direction, diacritics, dots, the absence of vowels and capital letters, connected letters without spacing between them, and a slanted calligraphic style. Another distinction of Arabic script is its use of contextual characters, also called positional variants. In the Arabic script, letters have four shapes depending on their position in the word: isolated (complete) form, initial form, medial form, and final form \cite{bilgin2023printed}. Accuracy of Arabic OCR models is negatively influenced by segmentation errors. To address the segmentation complexities and shape variations in Arabic, many systems adopt word-level OCR. Arabic script-specific preprocessing steps typically involve diacritic removal, dot normalization, and segmentation into words or characters. Models such as CNN, RNN, and LSTM have been used for Arabic OCR on a labeled Quranic text database \cite{mohd2021quranic}. Among these, GRU and LSTM-based models have shown high performance due to the sequential nature of the Arabic script.  To improve the Arabic OCR, postprocessing methods such as Word Beam Search decoding with dictionaries and auto-correction using BERT have been proposed \cite{krithiga2023ancient}.
 
\subsubsection{CHINESE SCRIPT}
The Chinese script, one of the world's oldest continuously used writing systems, dates back over 3,000 years to ancient oracle bone inscriptions \cite{liu2020optical}. Unlike alphabetic scripts, it is a logographic writing system where each character represents a word or a meaningful unit rather than individual sounds. The script is used primarily for Mandarin, Cantonese, and other Chinese dialects and has also historically influenced Japanese, Korean, and Vietnamese writing systems \cite{liang2025diversity}. Today, it is the official script of China, Taiwan, and Singapore, with variations in Simplified Chinese, which is used in mainland China and Singapore, and Traditional Chinese, which is used in Taiwan and Hong Kong.
Key features of the Chinese script include its top-to-bottom and left-to-right writing orientations, the absence of uppercase and lowercase distinctions, and diacritics. It consists of thousands of unique words made up of strokes and radicals, which provide clues for meaning and pronunciation. \cite{handel2019sinography}. Although complex, the Chinese script is one of the most visually attractive and historically significant writing systems globally. The massive Chinese character set, potentially containing over 50,000 characters in comprehensive dictionaries, is addressed through hierarchical encoding schemes and transformer architectures with attention mechanisms. To reduce computational complexity, OCR models employ multi-stage pipelines.
A large number of visually similar elements and stroke variations (1-30 strokes per character) also create computational challenges, tackled through multi-scale feature extraction, attention-based discriminative learning for subtle stroke differences, and synthetic data augmentation to improve character distinction.
The dual directionality with horizontal (left-to-right) or vertical (top-to-bottom, right-to-left) in Chinese text requires adaptive OCR systems. In literature, it is addressed through orientation-aware detection algorithms, bidirectional LSTMs for sequence modeling, and adaptive layout analysis modules in frameworks like EasyOCR and PaddleOCR \cite{kopitar2025review}.
\subsubsection{DEVANAGARI SCRIPT}
 In South Asia, the Devanagari script is one of the most widely used writing systems. It originates from the Brahmi script and has been used for over 1,500 years. It is the primary script for Hindi, Marathi, Nepali, and Sanskrit, among other languages \cite{arora2024devanagari}. 
The Devanagari script is an abugida, a writing system that differs from purely alphabetic, logographic, or syllabic scripts. In Devanagari, each consonant inherently carries a vowel sound, which can be modified or muted using diacritical marks. Unlike alphabetic scripts, where vowels and consonants are written separately, Devanagari combines them into a single unit \cite{gautam2020dataset}. This script is more structured, but it is also more complex than a simple alphabet. A distinctive feature of Devanagari is the horizontal line running along the top of the characters, which visually connects letters in a word. The script follows a left-to-right writing direction. It consists of vowel and consonant characters, diacritical marks, ligatures, and conjunct forms, where multiple consonants merge into a single shape. Digital text processing of Devanagari script is complicated since characters change form when combined, and words are written without explicit spacing between them \cite{singh2023feature}.

Consequently, segmentation of words, characters, and diacritical marks is a challenge for OCR on this script.
Segmentation methods are broadly divided into implicit (pure segmentation), explicit (recognition-based segmentation), and holistic (segmentation-free) methods. In Devanagari character segmentation, the header line is removed, and characters in words are divided into three zones: the upper zone, the middle zone, and the lower zone. However, the performance of OCR models is significantly reduced due to incorrect segmentation. Therefore, for the Devanagari script, custom segmentation algorithms for conjunct handling are proposed. For accurate Devanagari OCR hybrid models for OCR and post-processing for error correction are recommended \cite{manocha2021comparative}.  Given the inherently sequential and interconnected structure of Devanagari characters, Bidirectional LSTM (BiLSTM) is a more appropriate choice for accurately recognizing this script \cite{arora2024devanagari}.

\subsection{Key Challenges in Accurate OCR Systems (RQ3)}
\begin{figure}[h]
  \includegraphics[width=1\linewidth]{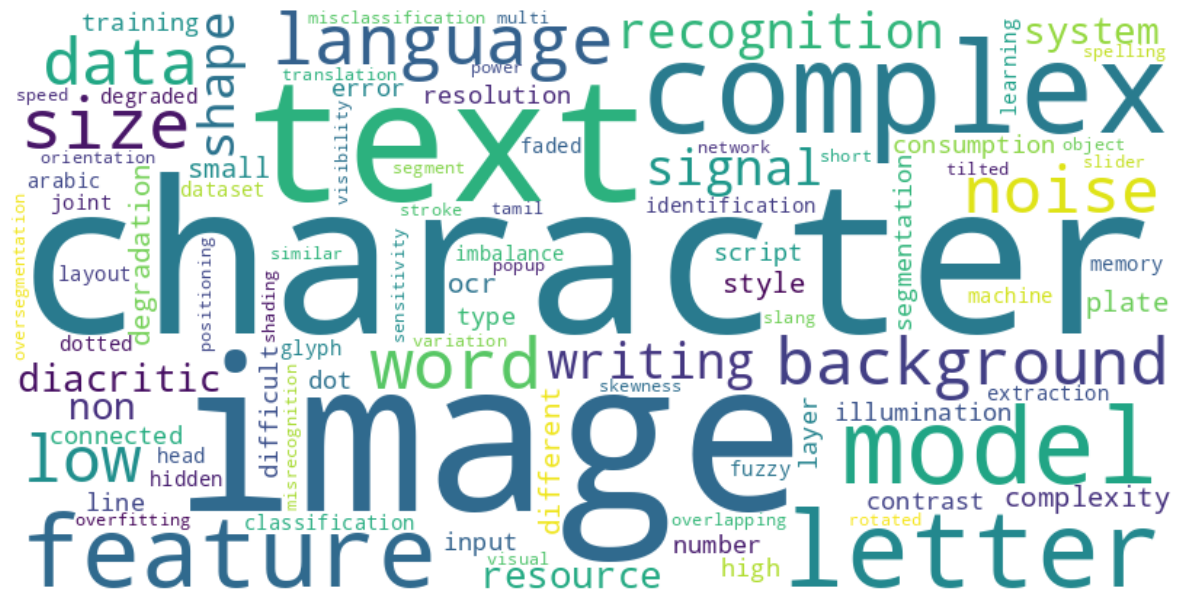}
    \caption{A word cloud of keywords extracted from literature phrases describing challenges.}
    \label{fig: Figure 11}
\end{figure}

A word cloud analysis was performed to identify key terminologies associated with the challenges and limitations reported in OCR research. Unlike conventional systematic reviews that primarily rely on titles, abstracts, and keyword-based filtering, this study adopted a full-text analysis. Consequently, more profound insights into the specific obstacles encountered in the field are provided as illustrated in \autoref{fig: Figure 11}. 

The most frequently reported challenges in OCR research emphasize the ongoing efforts to extract meaningful textual and visual features from complex, real-world images. Classification models, background noise, character shape and size variations, diacritics, low contrast, resolution limitations, and dataset scarcity are commonly discussed issues that cause persistent accuracy constraints. However, beyond these well-documented challenges, several less-explored but equally significant obstacles include glyph variations, distortion sensitivity, overlapping characters, skewed and tilted text, shading effects, data imbalance, positioning errors, short strokes, fragmented text segments, computational inefficiencies, overfitting, and punctuation mark recognition. These challenges bring up gaps in current OCR methodologies and point to deeper issues in script-specific complexities, language adaptability, and model generalization. The presence of these terms in word cloud feature the urgency to address these limitations with innovative solutions for advancing OCR technologies.

\begin{figure}[h]
    \centering
    \includegraphics[width=1\linewidth]{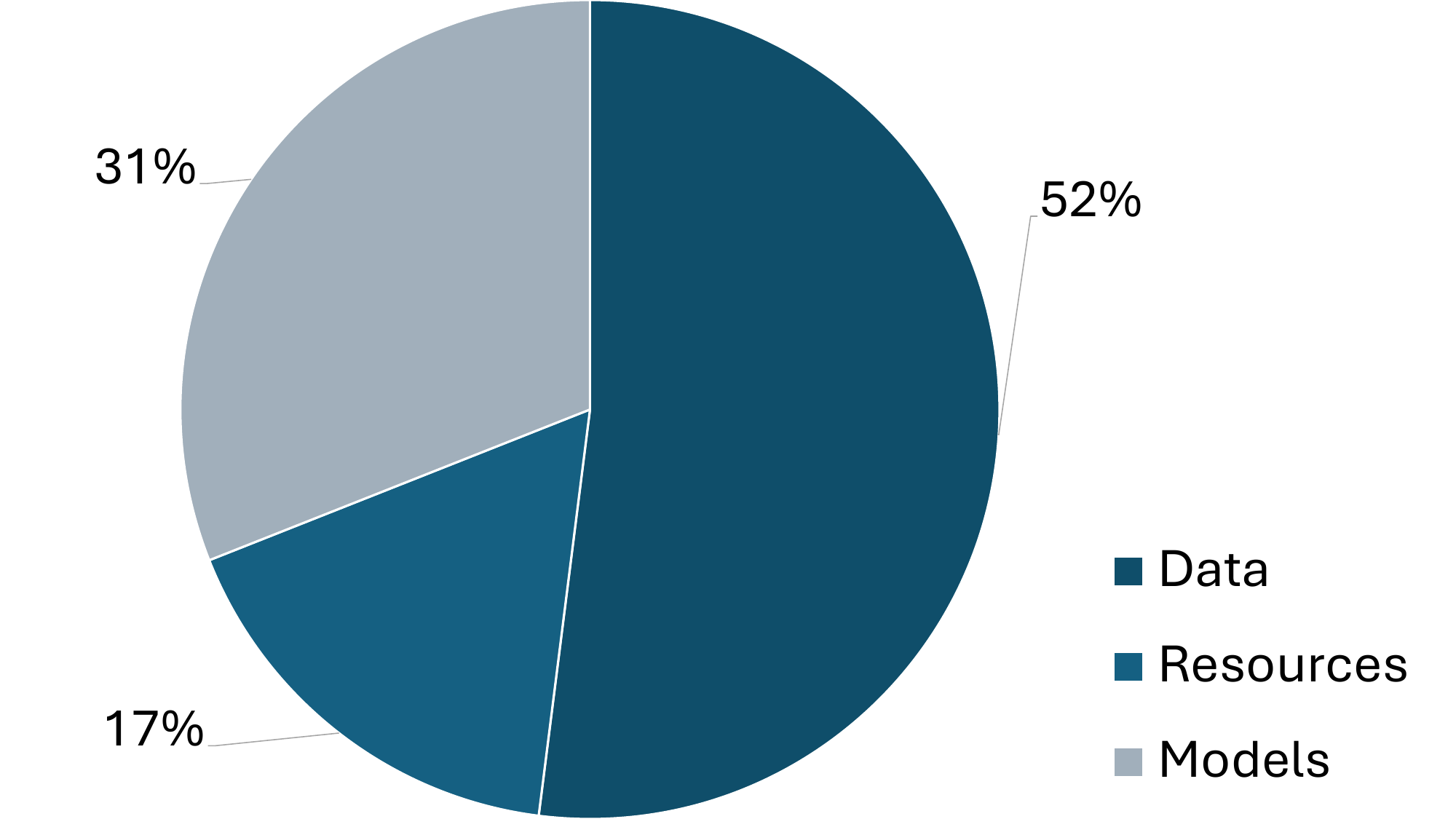}
    \caption{Main sources of challenges in OCR.}
    \label{fig: Figure 12}
\end{figure}

As observed, the studies on multi-disciplinary text extraction exhibited a variety of challenges and limitations that modern OCR models still face. The challenges reported in various contexts are broadly categorized into data quality issues, limitations of AI models (algorithmic challenges), and resource constraints. Explicitly mentioned limitations are outlined in \autoref{fig: Figure 12}.  

\begin{figure}[h]
    \centering
    \includegraphics[width=1\linewidth]{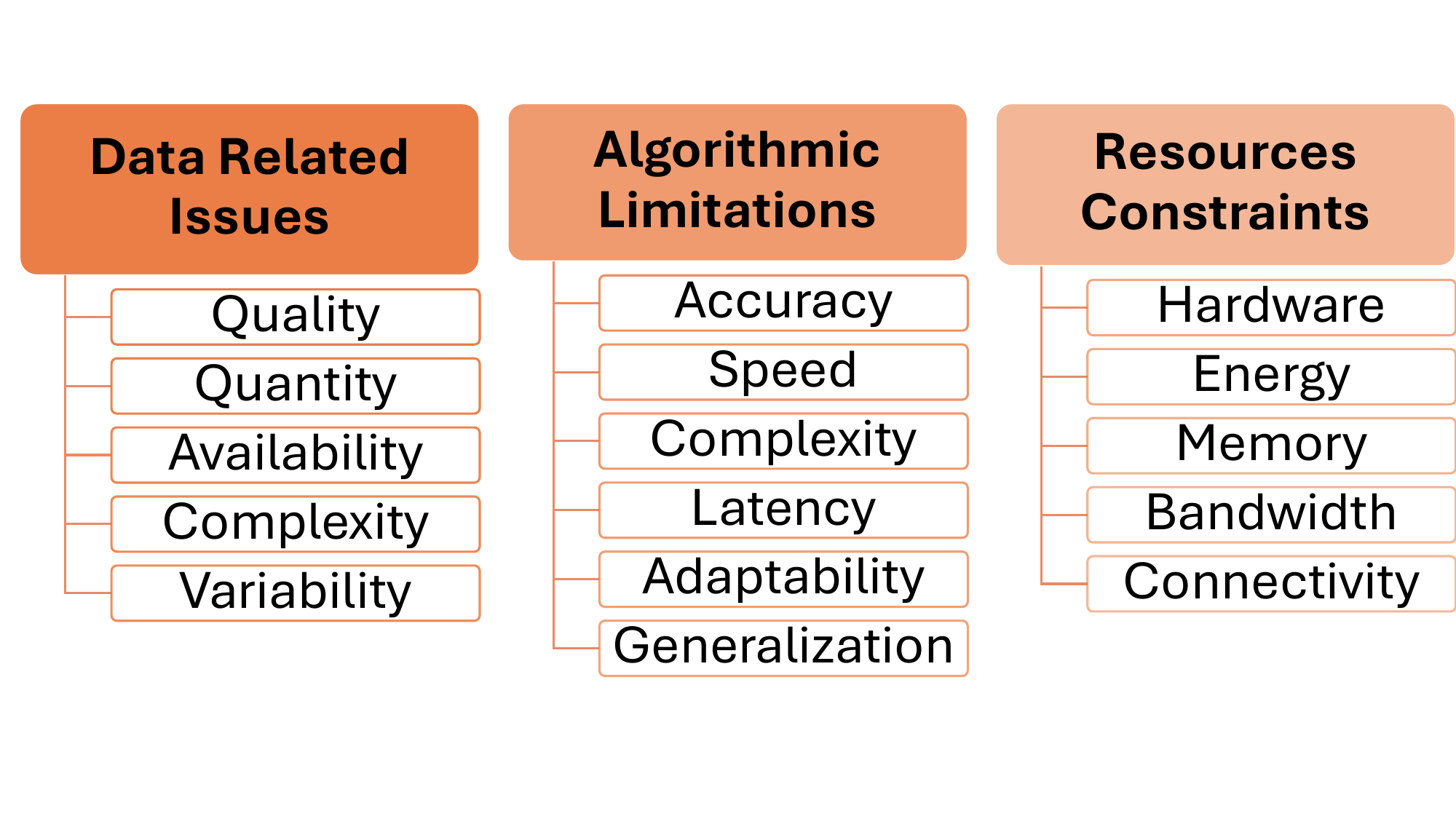}
    \vspace{-1cm}
    \caption{Key challenges and limitations in OCR.}
    \label{fig: Figure 13}
\end{figure}

Challenges reported in the literature are often problem-specific and closely linked to individual system designs. However, these challenges can be broadly categorized into key problem areas, as illustrated in \autoref{fig: Figure 13}. The identified challenges are grouped into data, model, and resource constraints. Since all main categories are interconnected while each category is wrapped around multiple subcategories. Therefore, the primary classification of challenges disregards their interconnected and multilayered nature.
To dig deeper, the prominent subcategories of challenges associated with the primary categories are shown in \autoref{fig: Figure 14}. Here, data-related issues focus on the quality, quantity, and availability of training data, directly impacting model performance. The complexity and variability of text, such as different fonts, handwriting styles, and distortions, add another layer of difficulty.
Model limitations represent the technical constraints of OCR systems, including accuracy, speed, and complexity. Besides, latency, adaptability to different text styles, and the ability to generalize through datasets and languages are significant hurdles, especially when dealing with noisy real-world text.
Finally, resource constraints are the practical barriers to deploying OCR systems. Hardware limitations include energy consumption, memory requirements, bandwidth, and connectivity issues. Resource constraints are particularly critical in edge computing or mobile applications, where processing power is limited.
Collectively, these categories illustrate the persistent challenges faced by OCR. Addressing these challenges requires identifying research gaps for continuous advancements in data processing, model efficiency, and resource optimization \cite{naseer2024investigating}. Since the challenges identified are potential avenues for future research, a thorough discussion of these challenges and solutions will be provided in a subsequent section.

The discussion is followed by targeted solutions for each challenge and general guidelines for enhancing OCR system performance in future research.
Data quality challenges include small sample sizes in datasets, insufficient for practical model training, and low-quality inputs, such as blurriness, noise, and low contrast. Complex scripts with diacritics and joint characters add to the difficulties, as do fonts, sizes, and shape variations that affect recognition accuracy. Moreover, structural problems such as overlapping text and connected characters make OCR tasks more challenging. Environmental factors such as changes in lighting and background complexity negatively affect text recognition in natural scenes.

Selecting and optimizing suitable classification algorithms requires domain-specific knowledge. The architecture of the model determines accuracy, speed, and adaptability. Complex models can enhance recognition but may increase computational costs and latency, which limit their real-time performance. A lack of optimal training data required by the model raises the risk of overfitting and the inability to generalize to new scripts and fonts. Moreover, similar shapes, distortions, or noise also complicate feature extraction.

Resource constraints inhibit the evolution of real-time and embedded OCR applications. Limitations in hardware, such as processing speed and memory, restrict model deployment in resource-constrained environments. High energy consumption and optimization issues arise from power efficiency constraints. The lack of publicly available OCR datasets limits training samples and reduces model generalization. Additionally, deep learning model complexity contributes to high latency and impacts real-time performance. 

Multilingual complexity is a growing challenge in OCR systems where script variations cause misclassifications and recognition errors.
Addressing these challenges  requires high-level data preprocessing techniques, optimized algorithmic strategies, and efficient resource management. The following sections propose targeted solutions for identified challenges and present general guidelines for developing robust OCR systems in the future.

The reviewed studies demonstrate substantial progress in OCR performance in several domains and uncover consistent shortcomings that limit the broader applicability and reliability of OCR technology. These limitations point to unaddressed critical gaps caused by script variability, domain-specific noise, and generalization on different databases. The following sections detail these research gaps obtained from empirical trends and common challenges observed in recent work.

\subsubsection{DATA-RELATED CHALLENGES}
\begin{figure}[ht]
    \centering
    \includegraphics[width=0.8\linewidth]{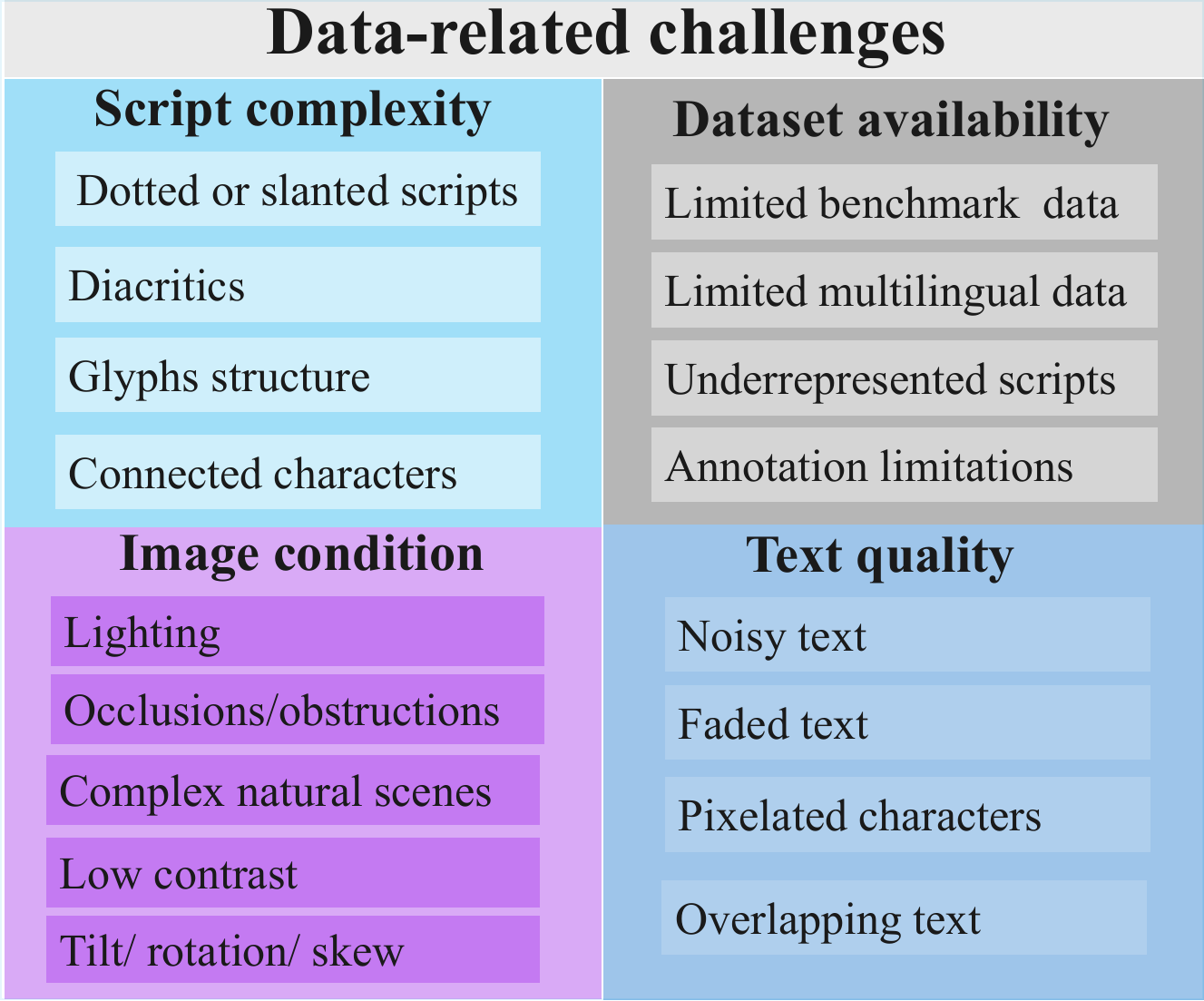}
    \caption{Data related challenges in OCR.}
    \label{fig: Figure 14}
\end{figure}

OCR models use supervised learning and primarily depend on extensive, high-quality labeled training data. A comprehensive, well-annotated, and balanced dataset with a sufficient number of examples is vital for achieving optimal training results. Many existing datasets today are limited in both size and scope \cite{tzoumpas2024data}. Uniform datasets are not available for different languages to equally train OCR models. Training OCR models on imbalanced datasets does not yield the necessary understanding for adequate generalization. Underfitting OCR models frequently fail when applied to real-world scenarios. Besides quantity, the quality of training datasets is an important element of OCR performance. Low-quality datasets propagate errors, resulting in poor performance even for strong models \cite{agrawal2024exploration}. Text data can be quite challenging for OCR systems to recognize accurately. Unlike typed English text in clean environments, real-world OCR tasks involve complex scripts with multiple strokes, diacritics, connected characters, and glyph structures that are difficult for models to learn without specialized data. Moreover, many datasets do not account for challenging conditions like noisy, low-resolution, or degraded documents. OCR models must contend with variations in illumination, low contrast, background clutter, and geometric distortions such as rotated, tilted, or skewed text. Factors reducing recognition accuracy are not universal for models to learn from datasets. Similarly, different scripts have unique annotation practices when it comes to segmentation. The absence of standardized annotations makes it difficult to train models, compare results, reproduce findings, and generalize to multilingual datasets. Continuous retraining is required to enhance performance and ensure reproducibility.
Furthermore, there has been limited attention given to documents with faded ink, worn-out characters, or overlapping and slanted text common in historical archives, handwritten records, and real-world imagery. Another limitation arises from domain-specific issues present in each dataset; for example, quality concerns differ between medical images and banking documents. Moreover, despite the increasing demand for multilingual OCR systems, there is a noticeable lack of high-quality datasets containing multiple scripts or mixed languages \cite{naseer2024investigating}. The common data-related issues, including availability, complexity, image condition, and text quality, are listed in Figure \ref{fig: Figure 14}.

\subsubsection{MODEL-SPECIFIC LIMITATIONS AND COMPUTATIONAL CONSTRAINTS}
Despite significant advancements in OCR technology, there are several limitations in recognizing complex text structures and scripts. Challenges associated with adopted models explicitly reported in the literature are provided in the \autoref{tab:ocr_summary} and described in the subsequent section. Key challenges existing OCR models face include the following:

\begin{table*}[h]  
\caption{Challenges associated with OCR models}
    \centering
    \renewcommand{\arraystretch}{1.3}
    \resizebox{\textwidth}{!}{  
    \begin{tabular}{p{3.5cm} p{4.5cm} p{6cm} p{2.5cm}}
        \toprule
        \textbf{Model} & \textbf{Application} & \textbf{Challenges Reported} & \textbf{References} \\
        \midrule
        CNN & Multiple studies on various languages (e.g., Arabic, Bangla, Telugu, Tamil, Persian, Urdu, Gujarati, Devanagari, Javanese and Swedish) & Small text size, blurry images, non-horizontal text, complex character shapes, compound characters, diacritics, connected characters, font variations, low-quality images, and overlapping text. & \cite{busa2022small,darwish2020enhanced, dessai2019deep, dipu2021bangla, khosrobeigi2022persian,drobac2020optical,ahlawat2020improved} \\  
        DNN & Studies on OCR for different languages and document types & High computational cost, large feature matrices, memory inefficiency, data imbalance, and difficulty in recognizing handwritten text. & \cite{wei2018improved,chaitanyaswami2023deep,zhang2022optical} \\  
        LSTM & Studies on handwritten OCR, license plate recognition, and document processing & Similar character shapes, joint characters, varying orientations, illumination conditions, font and background variations, and computational resource limitations.& \cite{brisinello2017improving,charan2022optical,arora2018optical,alamsyah2022autoencoder,gener2022gpgpu,wasan2022development,zou2020robust,andersson2022cognitive,rusli2021indonesian,kumaran2023text,vidakis2018facilitation,ramya2023number} \\  
        SVM & Printed Arabic OCR and Bangla OCR studies & Feature extraction challenges, low contrast images, and difficulty handling multiple font types. & \cite{gomes2017embedded,yamina2017printed,pervin2017feature} \\  
        ANN & Language-oriented OCR implementations for Arabic, Persian, and Nepali scripts & Image quality issues, font variations, low resolution, and complex background interference. &  \cite{rajkumar2022smart, afroge2016optical,singh2016optical,yap2021optical,sharma2017optical,chang2019optical}\\  
        Hybrid Models (CNN + LSTM, SVM + MLP, etc.) & Studies assessed ensemble approaches for OCR & Overfitting, increased training time, high memory requirements, and difficulty optimizing network layers. &\cite{zaryab2023optical,khosrobeigi2022persian, zhao2020improving, georgieva2020optical,sahlol2020handwritten, shi2022charformer,chandra2021end, eghbali2017font}  \\  
        GAN-based Models & Studies on OCR for character separation and text denoising & Complex backgrounds, overlapping text, and variations in stroke thickness. &  \cite{luna2022license,huang2021separating}\\  
        Transformer-based Models (TrOCR, ViT-OCR, etc.) & Studies on printed and handwritten OCR & Poor generalization for different languages, computational complexity, and model fine-tuning challenges. & \cite{azadbakht2022multipath,asadi2023farsi,zou2025tcmt} \\  
        \bottomrule
    \end{tabular}}
    \label{tab:ocr_summary}
\end{table*}

\begin{itemize}[left=0pt, labelsep=0.5em, align=left, itemsep=1pt, parsep=1pt]
\item Difficulty in handling overlapping, connected handwritten and cursive scripts.
\item High misclassification rates for visually similar characters, such as 'O' and '0' or '2' and 'z' lead to reduced recognition accuracy.
\item Ineffective segmentation techniques for complex layouts, handwritten scripts, and multi-column documents, resulting in fragmented or incorrect text extraction.
\item Deep learning models suffer from overfitting and inefficiencies when trained on small or unbalanced datasets.
\item Lack of adaptive OCR architectures to handle script, font, and language variations.
\end{itemize}

Deploying OCR systems on mobile, edge, and embedded devices often encounters computational challenges. A few frequently reported computational challenges are listed below:

\begin{itemize}[left=0pt, labelsep=0.5em, align=left, itemsep=1pt, parsep=1pt]
    \item Real-time OCR is impractical for resource-limited embedded systems due to high power consumption and processing overhead.
    \item Limited lightweight and optimized models are available for mobile and edge computing environments.
    \item Balancing accuracy, speed, and computational efficiency are challenging when optimizing deep learning architectures.
\end{itemize}

Understanding these collective challenges is essential for identifying research gaps. These gaps reveal where current models are deficient and what issues are holding back further progress. Tackling these issues is about making real advancements that push OCR models forward. The following section explores possible solutions and future directions to develop more adaptable and powerful OCR systems.

\subsection{OCR Enhancement Strategies (RQ3)}
\begin{figure*}[h]
    \centering \includegraphics[width=0.6\linewidth]{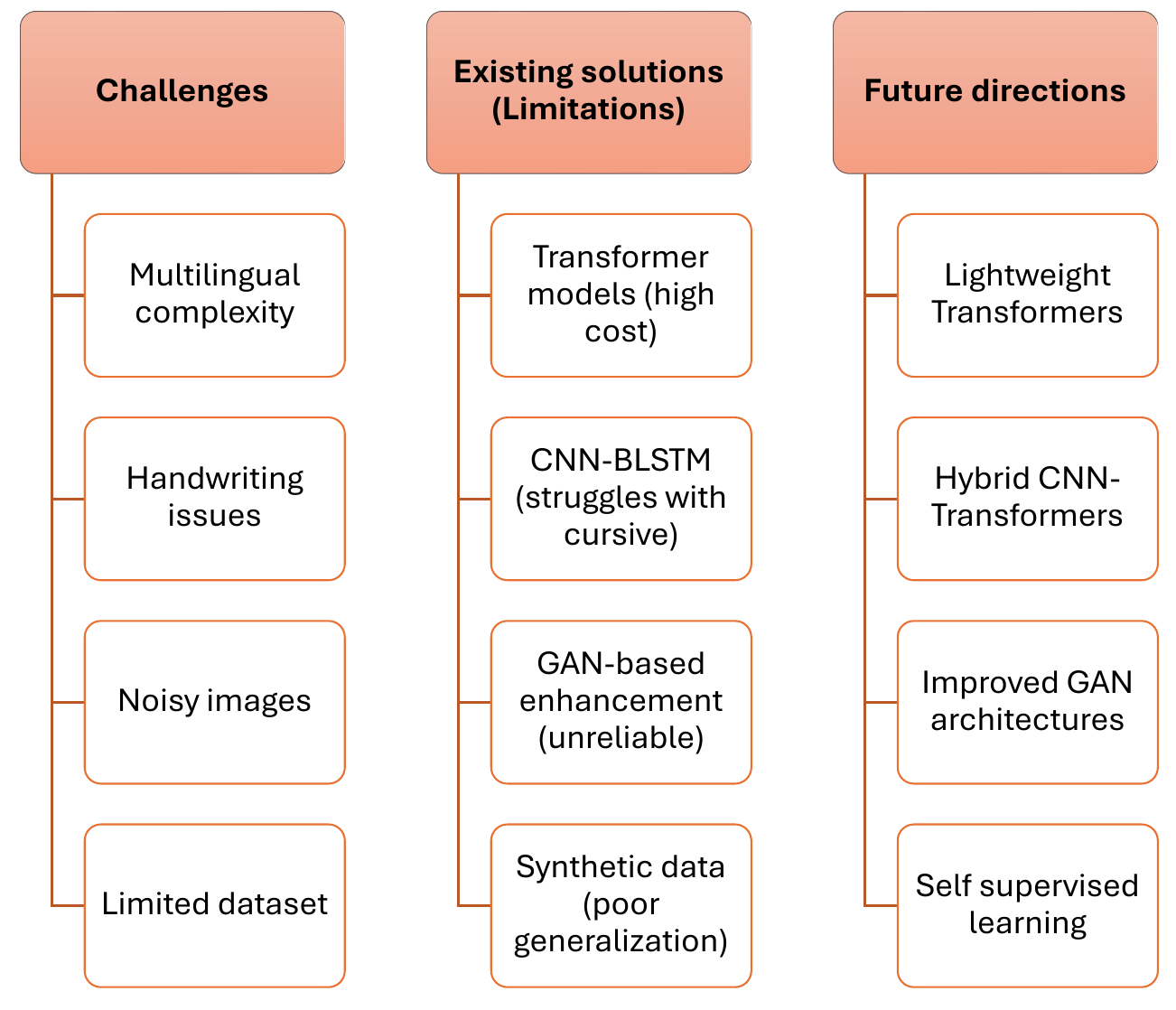}
    \caption{Frequently reported challenges, existing solutions, their limitations, and future directions to overcome these issues.}
    \label{fig: Figure 15}
\end{figure*}

Advanced strategies such as transformer-based architectures, self-supervised learning, and adaptive preprocessing methods are recommended for handling handwritten, complex, and cursive text recognition to enhance OCR performance further. Additionally, multi-modal learning, data augmentation, and GAN-based de-noising techniques should be integrated to mitigate challenges related to low resolution, distortions, and background interference \cite{pan2024lpsrgan}. Identified issues, existing solutions, the limitations associated with the existing methods, and future directions to overcome the challenges are in \autoref{fig: Figure 15}.

For specialized applications such as license plate and scene text recognition, domain-adaptive learning, improved segmentation, and multi-frame fusion methods can be explicitly applied to handle variations in lighting conditions and motion blur. Post-processing techniques, including NLP-driven context-aware correction, hybrid dictionary-based refinement, and ensemble OCR models, are suggested to enhance recognition accuracy. Furthermore, AutoML frameworks, hybrid deep learning models, and continual learning strategies should be explored for dynamic model adaptation without extensive manual tuning.
OCR is a multi-stage process involving data preprocessing, character recognition, and post-recognition refinement. Accuracy enhancement techniques are systematically integrated during all stages of OCR development. These techniques can be broadly categorized into three key strategies: data-centric, model-oriented, and output refinement.

\subsubsection{DATA-CENTRIC STRATEGIES}
A significant portion of OCR research focuses on improving input data quality to enhance recognition accuracy. Preprocessing techniques are widely employed to mitigate variations in lighting, resolution, and geometric distortions \cite{wei2018improved}. 

Data-centric strategies in OCR are developed to improve input image quality with preprocessing. It ultimately contributes to the accuracy and processing speed of OCR models. Unlike traditional model-centric approaches that refine OCR architectures, data-centric methods emphasize intelligent preprocessing, adaptive data augmentation, and efficient dataset curation to optimize recognition performance. Existing preprocessing techniques, such as binarization, noise reduction, contrast enhancement, and geometric correction \cite{wei2018improved}, have been widely used to address common challenges like poor lighting, distortions, and low resolution. However, preprocessing techniques are image- and case-specific and do not benefit images uniformly.

In the future, automated and flexible preprocessing pipelines can be developed to adjust data dynamically according to the characteristics of each image. Moreover, deep learning-based enhancement techniques, such as super-resolution reconstruction and adaptive denoising, can improve text clarity without relying on fixed parameters \cite{luna2022license}. Additionally, self-learning preprocessing frameworks, where models iteratively refine image quality based on OCR confidence scores, could enhance accuracy.

\subsubsection{MODEL-ORIENTED STRATEGIES}
Apart from input quality, OCR accuracy is ultimately determined by the ability of the recognition model to classify characters and words correctly. Model-oriented strategies focus on enhancing OCR systems through two primary approaches, model selection and optimization. first, the most suitable model is identified considering accuracy and computational efficiency. Then OCR models are fine-tuned by adjusting hyper-parameters, and employing suitable feature extraction techniques.

Model-oriented strategies in the literature still use domain knowledge for model selection and conventional optimization techniques for hyper-parameters tuning \cite{liao2022empirical}. We suggest automating these model selection and optimization strategies simultaneously using AutoML. It can streamline the selection of the most efficient OCR architectures and hyperparameters using AI.

\subsubsection{OUTPUT REFINEMENT STRATEGIES}
Despite improvements in preprocessing and model architecture, OCR outputs contain residual errors, commonly in complex text environments. Post-recognition refinement strategies correct misclassified characters and contextual inconsistencies to improve the reliability of the extracted text. Primary types of post-processing techniques include lexicon driven methods that correct OCR errors by referencing domain-specific dictionaries. Algorithm-based methods refine OCR predictions based on contextual understanding. Several computational models such as sequence-to-sequence (Seq2Seq) networks, topic-based language models, and probabilistic error correction algorithms are utilized. Besides, integrating domain-aware lexicons and language modeling techniques improve text recognition accuracy \cite{karthikeyan2021ocr}.
\subsubsection{AI INTEGRATED SYSTEMS}
\begin{figure}[h]
    \centering
    \includegraphics[width=1.08\linewidth]{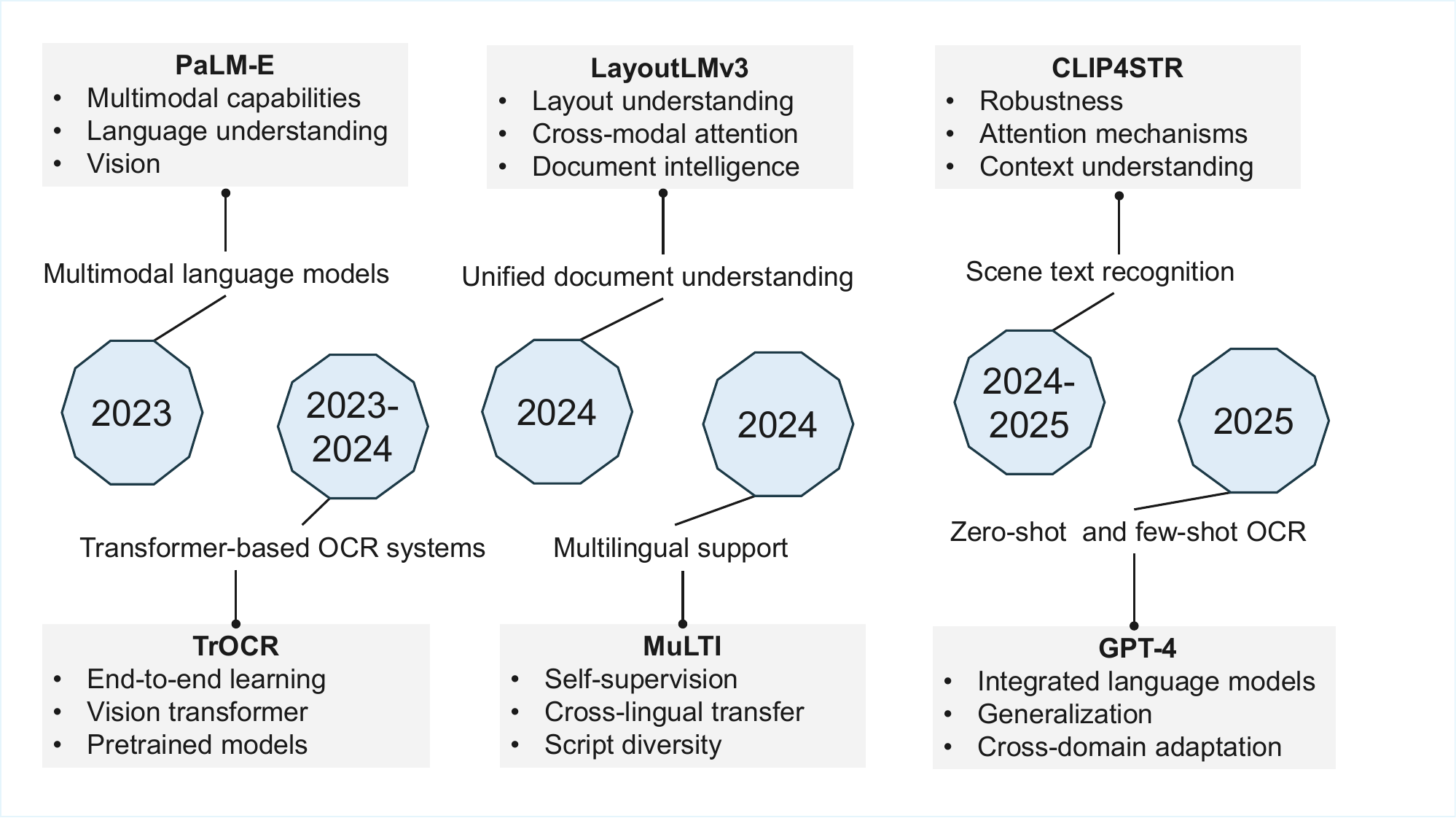}
    \vspace{-0.5cm}
    \caption{Recent advancements in integrated systems for OCR.}
    \label{fig: Figure 16}
\end{figure}

In recent years, OCR technology has moved from traditional image processing pipelines to more advanced, integrated systems that combine multiple areas of machine intelligence. Figure \ref{fig: Figure 16} shows the main stages of this development. The years indicate when each approach saw the most attention in literature and broad use, not necessarily when it was first introduced. Early OCR systems focused on low-level pixel operations, but over time, they grew into more capable systems that can interpret both visual and textual information. Their performance improved further as methods from computer vision, natural language processing (NLP), large language models (LLMs), and contextual reasoning were brought together. Introduced by Google,  PaLM-E demonstrated multimodal characteristics beyond conventional pipelines with embodied AI to interpret text within broader visual and semantic contexts \cite{gao2024point}. Along similar lines, transformer-based design of TrOCR established a standard approach of end-to-end learning, which mitigated the dependence on separate text detection and recognition. The same trend continued in LayoutLMv3, which combines visual, textual, and structural features through cross-modal attention. This combined OCR model is efficient for complex layouts like tables, forms, and mixed-content documents \cite{li2023trocr}. A few new scene-text detection models, such as CLIP4STR, PARSeq, and ABINet, apply vision-language pretraining and iterative correction for more reliable OCR in real-world conditions. Outside the main timeline, other important developments include DONUT, an image encode-text-decoder, which skips OCR entirely for document parsing \cite{sassioui2023visually}; Nougat, designed specifically for scientific documents; and BERT-based systems that extend document understanding through language modeling \cite{zhang2025imtlm}. Most recently, systems like GPT-4 Vision, Claude 3 Opus, and Gemini Pro Vision bring OCR into the zero-shot era for recognizing and interpreting text in unfamiliar domains and languages without task-specific training \cite{wang2024picture}. These models blur the line between OCR and general AI, and appear to be moving toward systems that not only extract text but also understand its visual, linguistic, and contextual framework.
While recent OCR models exhibit broader language understanding and some level of cross-domain capability, several improvements are needed in real-world applications. There is still no unified system that performs consistently well on heterogeneous document types and languages. Contextual understanding is limited for long documents. Most models also lack support for incremental learning and on-device adaptability. Future research should explore building multimodal OCR systems that can jointly process text, layout, and visual features. Addressing script diversity will require developing multilingual and multiscript models using self-supervised and cross-lingual learning. Further progress is needed in context-aware OCR using memory-augmented transformers. Models should be designed to generalize well in zero-shot settings while adapting to new domains and tasks. As OCR systems are widely adopted, protect user privacy with federated learning needs to be prioritized.

\subsection{Future Research Directions (RQ3)}

\subsubsection{ADVANCED LEARNING APPROACHES FOR L0W-RESOURCE LANGUAGES }
OCR on low-resource scripts is due to insufficient training data, which limits the ability of OCR models to generalize well. To address the dataset limitations, self-supervised learning (SSL) \cite{gui2024survey} and few-shot learning (FSL) \cite{gharoun2024meta} can be promising alternatives for extracting meaningful patterns from unlabeled text data. These techniques enhance recognition accuracy with minimal reliance on annotated datasets. Consequently, they reduce the need for extensive manual labeling. By integrating SSL and FSL, OCR systems can become more adaptable and resilient. In this way, text recognition can be more accessible to languages with scarce digital resources.

\subsubsection{OPTIMIZED ARCHITECTURES FOR REAL-TIME APPLICATIONS}
While transformers have significantly advanced OCR performance, their computational cost is a key limitation for real-time deployment. Research on efficient transformer variants such as MobileViT and TinyBERT can improve model scalability without compromising accuracy \cite{azadbakht2022multipath}. Quantization, knowledge distillation, and pruning techniques can further reduce model size and inference time \cite{kim2023quantization}, making OCR viable for mobile and embedded systems.
\begin{table*}[!b]
    \centering
    \caption{Comparison of traditional image enhancement techniques and GAN-based methods. (\cmark = Supported, \xmark = Not supported, \partialmark = Limited support)}
    \label{tab:enhancement_comparison}
    \footnotesize
    \renewcommand{\arraystretch}{1.1}
    \begin{tabular}{>{\raggedright\arraybackslash}p{3cm}*{8}{>{\centering\arraybackslash}p{1.3cm}}}
        \toprule
        \textbf{Enhancement Capability} & \textbf{Bilateral Filter} & \textbf{CLAHE} & \textbf{Otsu Thresholding} & \textbf{Gaussian Blur} & \textbf{Unsharp Mask} & \textbf{Morph. Ops} & \textbf{Wavelet Transf.} & \textbf{GAN Methods} \\
        \midrule
        Noise Reduction              & \cmark & \xmark & \xmark & \partialmark & \xmark & \xmark & \cmark & \cmark \\
        Contrast Enhancement         & \xmark & \cmark & \xmark & \partialmark & \xmark & \xmark & \xmark & \cmark \\
        Resolution Enhancement       & \xmark & \xmark & \xmark & \xmark       & \xmark & \xmark & \xmark & \cmark \\
        Blur Removal                 & \xmark & \xmark & \xmark & \partialmark & \cmark & \xmark & \xmark & \cmark \\
        Artifact Removal             & \xmark & \xmark & \xmark & \xmark       & \xmark & \xmark & \xmark & \cmark \\
        Illumination Correction      & \xmark & \cmark & \xmark & \xmark       & \xmark & \xmark & \xmark & \cmark \\
        Data Generation              & \xmark & \xmark & \xmark & \xmark       & \xmark & \xmark & \xmark & \cmark \\
        Multi-degradation Handling   & \xmark & \xmark & \xmark & \xmark       & \xmark & \xmark & \xmark & \cmark \\
        Semantic Understanding       & \xmark & \xmark & \xmark & \xmark       & \xmark & \xmark & \xmark & \cmark \\
        Adaptive Processing          & \xmark & \xmark & \xmark & \xmark       & \xmark & \xmark & \xmark & \cmark \\
        \bottomrule
    \end{tabular}
\end{table*}

\subsubsection{ADVANCED IMAGE ENHANCEMENT TECHNIQUES}
Noisy and low-resolution images are one of the key obstacles to OCR accuracy. Future research should focus on GANs for document enhancement. Super-resolution GANs (SRGAN, DESRGAN) can reconstruct degraded text, mitigate background noise, and enhance text clarity to improve recognition performance in challenging real-world environments.

An important prerequisite for achieving high OCR accuracy is the completeness and legibility of both training and inference data \cite{neudecker2021survey}. In many real-world scenarios, parts of the textual information may be missing or corrupted. Unlike traditional enhancement filters that only modify existing pixels, GANs can generate entirely new image content based on learned patterns \cite{ibrahim2022license}. Through adversarial training, GANs produce high-quality reconstructions with lifelike features that are sufficiently realistic to fool the discriminator network \cite{kopitar2025review}.

With semantic understanding, GANs are capable of reconstructing contextually appropriate character structures. They address both local character-level missing information and global document-level quality issues simultaneously \cite{li2025matching}. In contrast to conventional methods, GANs learn the data distribution and hallucinate plausible structures in place of missing or degraded regions. Consequently, GAN-based methods can restore broken license plates \cite{pan2024lpsrgan} obscured by dirt or weathering \cite{pattanaik2023enhancement}, reconnect incomplete pen strokes in handwritten text \cite{ayyoob2024stroke}, recover damaged or faded characters in historical documents \cite{souibgui2020gan, saha2024npix2cpix}, infer obscured portions of text hidden by ink smearing (ink blots) \cite{alkendi2024unsupervised,kodym2022tg} and overlapping objects in scene text images \cite{cheema2024novel}. For challenging images with noise or missing information, GAN-based preprocessing for simultaneous data cleaning, repairing, filling, and reconstruction instead of simple image enhancement is expected to overcome the limitations of traditional preprocessing techniques. A comprehensive comparison of GAN with traditional popular data-centric approaches, including Bilateral filter, Contrast-limited adaptive histogram equalization (CLAHE) \cite{musa2018review}, Otsu thresholding \cite{amiriebrahimabadi2024comprehensive}, Gaussian blur \cite{liu2020estimating}, Unsharp mask \cite{shine2023approach}, Morphological operations (Morphological Ops), and Wavelet transform is provided in Table \ref{tab:enhancement_comparison}.

\section{Conclusion}\label{sec:conclusion}
The evolution of AI models for Optical Character Recognition (OCR) has led to remarkable improvements in text recognition accuracy. Yet, significant challenges persist in handling complex scripts, handwritten text, and real-world noisy environments. This systematic review analyzed 97 research articles from seven major scholarly databases to present a structured examination of OCR progress over the past decade. It investigated the evolution of OCR models, the expansion of application domains, the increasing complexity of dataset types, and the continuous improvements in accuracy. Through a PRISMA-based methodology, this study examined the transition from traditional machine learning techniques (KNN, RF, SVM) to deep learning-based architectures, including CNNs, LSTMs, GANs, and Transformers, which have significantly enhanced OCR performance.
This review explores advancements and highlights critical gaps in OCR research. The main challenges fall into three areas: data limitations, algorithmic constraints, and computational demands.

To address these issues, we propose targeted solutions. Techniques such as transfer learning and self-supervised learning can enhance model generalization, while multimodal AI and integrated data sources help mitigate data scarcity. For improved accuracy, the use of advanced pre-trained models is recommended to refine OCR outputs at the character level. This survey provides valuable insights for both industry and academia. The identified research gaps and suggested solutions are expected to accelerate the development of fast, reliable, and multilingual AI-driven OCR systems.

\section*{Acknowledgment}
The authors extend their gratitude to UTM for providing invaluable support.

\bibliographystyle{IEEEtran}
\bibliography{cas-refs.bib}

\vspace{-2em}
\begin{IEEEbiography}[{\includegraphics[width=1.12in,height=1.3in,clip,keepaspectratio]{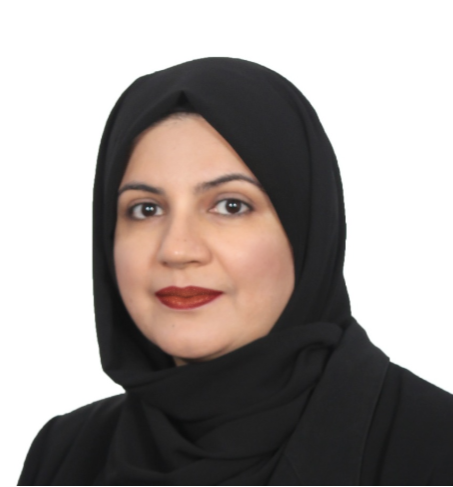}}]{Nuzhat Khan}
is a Postdoctoral Research Fellow at the VLSI and Embedded Computing Architecture Design Laboratory (VeCAD Lab), Universiti Teknologi Malaysia (UTM). She completed her Ph.D. in Environmental Technology at Universiti Sains Malaysia (USM) in 2023. She earned her M.S. in Information Technology from Balochistan University of Information Technology, Engineering, and Management Sciences (BUITEMS), Pakistan, in 2019 under the Higher Education Commission (HEC) Scholarship. Her research focuses on optical character recognition (OCR), image processing, natural language processing (NLP), artificial intelligence (AI), machine learning, deep learning, applied and statistical linguistics, and precision agriculture. She has authored multiple research papers, completed projects, and holds patents in these domains. She is a technical committee member for the ICEST 2025 conference. She serves as a reviewer for several reputed journals, including Pattern Recognition, Genetic Resources and Crop Evolution, Artificial Intelligence Review, Scientific Reports, Theoretical and Applied Climatology, Journal of Oil Palm Research, Discover Applied Sciences, Cluster Computing, and the Journal of Big Data, among others. 
\end{IEEEbiography}


\begin{IEEEbiography}[{\includegraphics[width=1.1in,height=1.30in,clip,keepaspectratio]{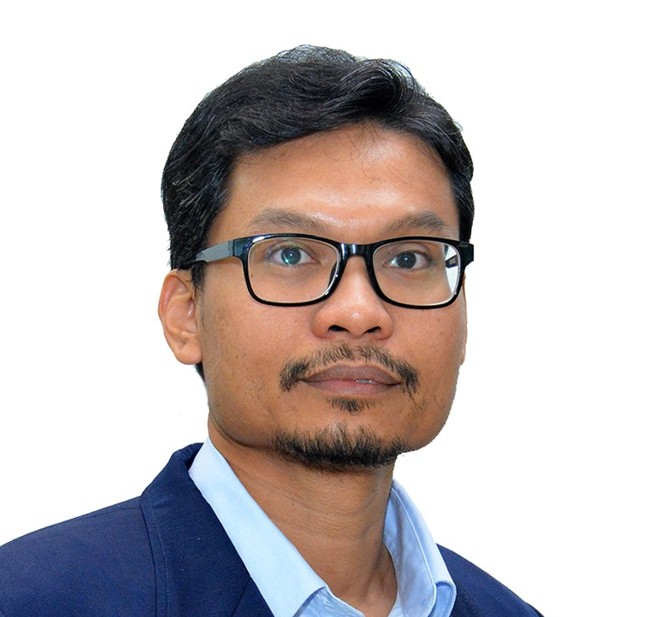}}]{Ab Al-Hadi Ab Rahman}
PhD, SMIEEE, MIET, CEng, PEng, is the Head of the Research Group at the VLSI and Embedded Computing Architecture Design (VeCAD) Laboratory, Faculty of Electrical Engineering, Universiti Teknologi Malaysia (UTM). He is also an Associate Professor in the Department of Electronic and Computer Engineering at UTM. Dr. Ab Al-Hadi holds professional affiliations as a Senior Member of the IEEE (SMIEEE), a Member of the Institution of Engineering and Technology (MIET), a Chartered Engineer (CEng), and a Professional Engineer (PEng). His research interests include VLSI design, embedded systems, computer architecture, and high-performance computing. He has published extensively in high-impact journals and conferences. He leads several research projects focused on advanced computing and embedded system technologies. He also holds two patents in the related area.
\end{IEEEbiography}


\begin{IEEEbiography}[{\includegraphics[width=1.1in,height=1.3in,clip,keepaspectratio]{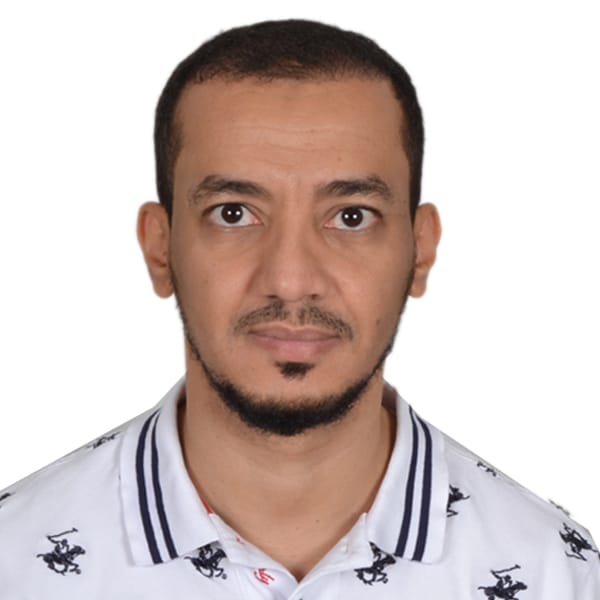}}]
{Ibrahim Yousef Alshareef} is pursuing his Ph.D. in Artificial Intelligence at Universiti Teknologi Malaysia (UTM). He holds a bachelor's degree in industrial electronics and automatic control from the College of Electronics and Communications and a Master's in Embedded Software Engineering from Gannon University, USA. His research interests include deep neural networks, electronics, and embedded systems, focusing on optimizing computational efficiency and performance in artificial intelligence applications.\end{IEEEbiography}
\vspace{-1em} 

\begin{IEEEbiography}[{\includegraphics[width=1.1in,height=1.3in,clip,keepaspectratio]{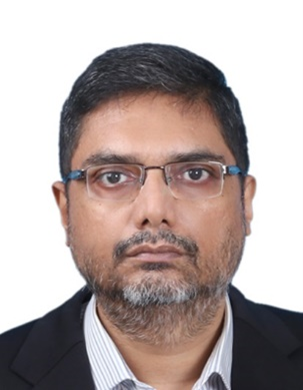}}]{Shahriyar Masud Rizvi} is an Associate Professor in the Department of Electrical and Electronic Engineering and a Deputy Director at the Center for VLSI and Embedded Systems, Dr. Anwarul Abedin Institute of Innovation at the American International University-Bangladesh (AIUB). He received his BS and MS degrees in Electrical Engineering (EE) in 2000 and 2002, respectively, from the University of Wyoming, WY, USA. He obtained his Ph.D. in EE in 2023 from the Universiti Teknologi Malaysia (UTM), Johor Bahru, Malaysia. His research interests include low-complexity and energy-efficient deep learning, optical character recognition (OCR), synthesis methodologies for behavioral FSMs modeled in SystemVerilog/VHDL, and hardware/software co-simulation for FPGA-based image processing. He also serves as a reviewer for the Scopus-indexed AIUB Journal of Science and Engineering (AJSE) and numerous international conferences.
\end{IEEEbiography}
\begin{IEEEbiography}[{\includegraphics[width=1.1in,height=1.3in,clip,keepaspectratio]{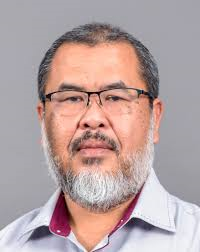}}]{MUHAMMAD NADZIR MARSONO} is an Electronics and Computer Engineering Professor at Universiti Teknologi Malaysia (UTM). He serves as Deputy Dean (Academic and Student Affairs) at the Faculty of Electrical Engineering. He is also a member of the VeCAD research group. His research focuses on domain-specific computing architecture, VLSI/SoC design, and embedded systems. His interests span digital system design, computer architecture, reconfigurable computing, multicore/manycore SoCs, NoCs, network algorithmics, network processing, and software-defined networks. He holds a PhD from the University of Victoria, Canada, and an MEng degree from UTM.
\end{IEEEbiography}
\begin{IEEEbiography}[{\includegraphics[width=1.1in,height=1.3in,clip,keepaspectratio]{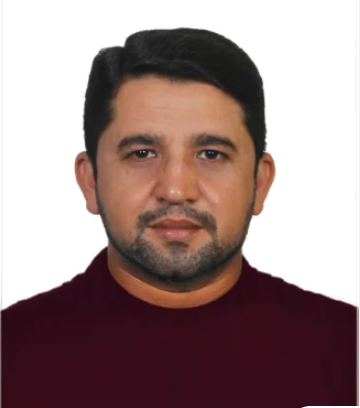}}]{Muhammad Paend Bakht}
received the Ph.D. degree in electrical engineering from the Universiti Teknologi Malaysia (UTM). He is currently an Assistant Professor in the Department of Electrical Engineering at Balochistan University of Information Technology, Engineering and Management Sciences (BUITEMS), Quetta, Pakistan. He has also gained postdoctoral research experience at Universiti Tun Hussein Onn Malaysia (UTHM). He has authored several research articles in reputable journals and presented his research ideas at international conferences. His research interests include optimization and energy management of grid-connected renewable energy systems, Artificial Intelligence, and predictive modeling. His current research at Teagasc, Ireland, focuses on the integration of renewable energy technologies in agriculture.
\end{IEEEbiography}
\begin{IEEEbiography}[{\includegraphics[width=1.1in,height=1.3in,clip,keepaspectratio]{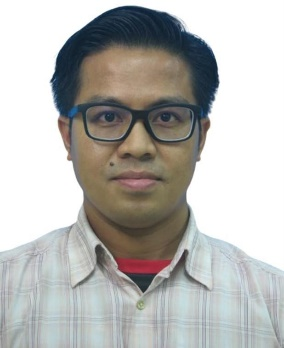}}]{MOHD SHAHRIZAL RUSLI} received his PhD, MEng. and BEng. in electrical engineering from Universiti Teknologi Malaysia (UTM), Malaysia. During his doctoral study, he attended the University of California Irvine, USA, and the University of Catania, Italy, as a research scholar. His research interests are network-on-chip, computer architecture, machine learning accelerators and power management. He has published in numerous journals, proceedings, and book chapters. From 2021 to 2022, he was a SoC design engineer at Mediatek Singapore, working on interconnect design and timing closure. He is a senior lecturer of Electronic and Computer Engineering, UTM, and a researcher for VLSI and Embedded Computing Architecture Design (VeCAD) Research Group.
\end{IEEEbiography}
\begin{IEEEbiography}[{\includegraphics[width=1.1in,height=1.3in,clip,keepaspectratio]{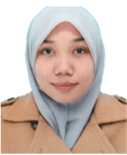}}]{SHAHIDATUL SADIAH} 
(Member, IEEE) received a B.Eng. degree in communication network engineering and an M.Eng. degree in electronic and information system engineering from Okayama University, Japan, in 2013 and 2015, respectively, and a Ph.D. degree in information engineering from Hiroshima University, Japan, in 2018. She has been a senior lecturer with the Department of Electronics and Computer Engineering, Universiti Teknologi Malaysia, since January 2019. Her research interests include cryptography, information security, digital system design, and computer architecture.

\end{IEEEbiography}

\EOD
\end{document}